\documentclass[runningheads]{llncs}
\usepackage[T1]{fontenc}
\usepackage{graphicx}
\usepackage{color}
\usepackage{soul} %For highlighting, used by Bastien

\usepackage{url}

\usepackage{times}
\usepackage{latexsym}

\usepackage[T1]{fontenc}
\usepackage[utf8]{inputenc}

\usepackage{microtype}

\usepackage{inconsolata}

\usepackage{graphicx}

\usepackage{amssymb} % Nécessaire pour \mathbb
\usepackage{amsmath}
\usepackage{amsmath,amsfonts,bm}

\def\eqref#1{equation~\ref{#1}}
\def\1{\bm{1}}

\DeclareMathAlphabet{\mathsfit}{\encodingdefault}{\sfdefault}{m}{sl}
\SetMathAlphabet{\mathsfit}{bold}{\encodingdefault}{\sfdefault}{bx}{n}

\DeclareMathOperator*{\argmin}{arg\,min}

\newcommand{\fh}{\hat{f}}
\usepackage{hyperref}
\newcommand{\hidden}[1]{}

\usepackage{url}
\usepackage{graphicx}

\usepackage[most]{tcolorbox}
\newtheorem{obs}{Observation}
\begin{document}
\title{Re-examining learning linear functions in context}

%\title{Contribution Title}
\author{Omar Naim\inst{1,2}\orcidID{0009-0006-2924-0421} \and Guilhem Fouilhé\inst{1,2}\orcidID{0009-0002-7108-9930} \and Nicholas Asher\inst{1,3}\orcidID{0000-0002-7689-8246}}

\institute{Institut de Recherche en Informatique de Toulouse, IRIT France \and Université de Toulouse, France \and CNRS, France \\ \email{ omar.naim.docs@gmail.com \\
\{guilhem.fouilhe-lafforgue, nicholas.asher\}@irit.fr }}

%\institute{Princeton University, Princeton NJ 08544, USA \and
%Springer Heidelberg, Tiergartenstr. 17, 69121 Heidelberg, Germany
\maketitle              % typeset the header of the contribution
\begin{abstract}

 We explore in-context learning (ICL), a popular paradigm for inference with Large Language Models (LLMs), in a controlled experimental setup using synthetic training data.  Using a range of small transformer models trained from scratch, we focus on a mathematical task with simple yet precise prompts: learning a linear function $f$ from a sequence of inputs $x_i$ and their corresponding function values $f(x_i)$.  
 Our findings challenge the prevailing narrative that transformers adopt algorithmic approaches like linear regression to in-context learn (ICL) a linear function.  We observe that all models have ``boundary values'' that limit generalizability.  While we can extend boundary values with training distributions over a wider range, we lose the precision of models trained on distributions with more restricted ranges.  Thus, we see a dilemma for ICL at least in some tasks: either models will lack generalizability or precision.

\end{abstract}
\section{Introduction}

Large language models (LLMs) exhibit a remarkable capability to learn by analogy, known as in-context learning (ICL) \cite{brown:etal:2020}.  %ICL enables the model to learn a task from a prompt with instructions and a few examples at inference time, without any adjustment of the model's parameters from pretraining. 
This approach is especially useful for adapting a model to a task with very few resources at inference time.  Despite several theoretical reconstructions of ICL, there have been few studies on exactly how ICL works in practice.  Are models actually learning to reason or are they simply memorizing training data, as \cite{gsm-symbolic} have suggested for other tasks? %This is a question, you need a question mark 
We address this gap by focusing on a simple mathematical task, the in-context learning  %"in-context learning" instead
of one-dimensional linear functions, where the training regime and data are straightforward and clean. This allows us to have greater control over the data, enabling a more effective analysis of the model's ICL behavior.

We distinguish between two notions of "learning a function".  The first, assumed by the literature since \cite{garg:etal:2022}, says that a model has learned a function if, when sampling values and coefficients of linear functions at test time from the same distribution used in training the model yields close to $0$ expected error.  Our models can perform ICL for linear functions in this sense, which we call {\em ICL}$_1$.  On the other hand, another notion, call it %souds very familiar, we could say "let's call it" to add better readibility but it stays familiar or we could write "which we call ICL2" but there is a répétition with the previous "which we call ICL1", so why not changing on of them by "which we refer to as"
{\em ICL}$_2$, might say that learning a linear function requires specifying the parameters $a,b$, where $f(x) = ax +b$, from data points in the prompt, something which is clearly possible using a variety of techniques like linear regression.  \cite{akyurek:etal:2022,vonoswald:etal:2023,ahn2023transformers,mahankali2023one} claimed that transformers use such techniques and so exhibit ICL$_2$. % And so *does* ICL2, I think you are missing a word here
However, we give strong empirical evidence that none of our transformer models use such techniques or are performing ICL$_2$ % can we really use ICL as a verb ? I would say "do ICL2 on linear function"
of linear functions. 

The reason our models cannot ICL$_2$ is that, across all training distributions, they have ``boundary values"  $B^{-}, B^{+}$ that constrain their predictions. When the inputs $x_i$ and the function $f$ parameters are chosen such that $f(x_i) >> B^{+}$ or $<< B^{-}$,  model performance degrades substantially.  We investigate the conditions under which these boundary values arise. %Our experiments indicate that attention layers are both necessary and sufficient for the boundary values.  
These specific values, frequently near the maximum and minimum observed during training, indicate that the model has likely memorized them.  But while boundary values are defined by training,  the models' degraded predictions do not come from the training data but are rather a sort of reasoning-based hallucination.  %Uniform distributions over a given interval gave a better idea of limits on generalizability for the models; in particular, all models have `  % {\color {magenta} The largest models profited from prompts ordered according to the natural order on $\mathbb{R}$, though smaller models did not.}  

 %attention + MLP layer to ICL ${\cal L}$. This enables them to learn a projection we describe in Section 5 from ``nearby" sequences of points in the training data to find the target function. %; if one means that the model can approximate the plot of a linear function arbitrarily well over a limited interval---what we call {\em ICL ${\cal L}$ plots over definite intervals}---then we answer "yes".  Making this point clearly helps us understand ICL better.  

We studied ICL in over 30 models, ranging from transformer architectures with a single layer and one attention head (AH) to models with 12 layers and 8 AHs, as well as small attention-only models \cite{olsson:etal:2022}. To train our models, we use % use
sequences generated by functions $f(x) = ax +b$ where %", where"
$a, b$ are sampled from a distribution over function parameters and $x$ is drawn from a distribution over inputs.  The model learns an algorithm on that training data, and we then evaluate its predictions on a test distribution at inference time.
Since ICL depends on a model's pretraining, such an investigation requires training from scratch, and that has constrained our study to relatively small transformer architectures.

\section{Related Work}
\cite{brown:etal:2020} introduced ICL as a paradigm where the model learn at inference from the prompt by analogy, with no changes to the parameters set during training.
Despite its promise and popularity, this phenomenon is still not fully explored.  
%\cite{brown:etal:2020} introduced ``in-context learning".  %, there has been considerable research to understand and explain this pattern.%indicating that ICL is possible because of a sort of gradient ``ascent'' \guilhem{Why do you specifically say "ascent" ? it's just a matter of objective function and vonoswald says descent in their title ; and akyurek in their abstract also descent. I would suggest "investigating the mechanisms of this learning algorithm \\cite..."  Or something like this. Note that you below talk about ICL and optimisation more in depth so it's okay to be general here. } \cite{akyurek:etal:2022,vonoswald:etal:2023}.
\cite{dong2022survey} surveys the successes and challenges for ICL and argue that so far, research has only studied simple tasks like learning linear or %remove this simple as it is a repetition and it feels like you are cheating on your comparison "it's as simple as something simple"...
Boolean functions.

\cite{garg:etal:2022} showed that a transformer trained from scratch (GPT-2 \cite{radford2019language} with an embedding size of 256) performed in-context learning of linear functions given identical train and test Gaussian distributions. %do ICL linear functions, but in their experiments the test distribution was the same as the train and  mainly for values $a,b \in \mathcal{N}(0,1)$.  %This means that the majority of the values of $a$ and $b$ far from $[-1,1]$ were not examined at all.  We based our code on theirs.% and perform in-context learning of linear functions with models (l,h,64) with $l \in \{1-6\}$, and AH $h \in \{1,2,4\}$, but we trained also bigger models for our experiments such GPT2 (12,8,256).

Further research has offered several reconstructions for how ICL might work for the class of linear functions ${\cal L}$ in transformers. \cite{akyurek:etal:2022,vonoswald:etal:2023,ahn2023transformers,mahankali2023one} provided a construction to show that transformers could learn via gradient descent to do linear regression in ICL, which would solve the task. \cite{fu2023transformers} showed that transformers could perform ICL %again here, icl isn't a verb
in virtue of using higher-order optimization techniques. \cite{xie2021explanation,wu2023many,zhang2023and,panwar2023context} argued that ICL follows from Bayesian principles. \cite{bai2024transformers} show that transformers can under certain assumptions implement % I think what you wanted to write is "show that, under certain assumptions, transformers can implement [...]"
many algorithms with near-optimal predictive power on various in-context data distributions.   Given \cite{perez:etal:2021}'s result that full transformers with linear attention are Turing complete, however, such theoretical demonstrations are not surprising. %why ? :') ", given the ... "
Whereas most prior research has concentrated on what transformer models {\em can} or {\em could} do on this task, we examine how ICL works in practice under different training and testing distributions in order to establish what transformers {\em actually} do.% in ICL 1 dimensional linear functions

\cite{xie2021explanation,zhang:etal:2024}  show that when the training and inference distributions are shifted,  ICL performance degrades. This work is closely aligned with our approach, as is the work of %I will be honest I am not even sure it's gramatically correct, "This work is closer to our own, as is that of \cite"
\cite{giannou:etal:2024}.  However, \cite{giannou:etal:2024,zhang:etal:2024} make important modifications to transformer architectures by working with linear attention, whereas we look at attention layers as they are more commonly implemented with softmax. In addition, \cite{zhang:etal:2024} uses a new kind of optimization or training with gradients and a special fixed initial point. This means that their architecture and training are quite different from what normally happens with transformers; they are interested in getting a revised transformer-like model to learn linear functions, while we want to find out whether existing transformers learn linear functions or something else.  As we detail below, the results for the architectures of \cite{zhang:etal:2024,giannou:etal:2024} are quite different from those we have for standard transformers. %Unlike either of these papers, we show that prompts with lengths longer than seen during training yield chaotic behavior. 
Finally, our proposed mathematical model for ICL differs from any of the proposed methods. 

\cite{bhattamishra2023understanding}  trained small GPT-2-like models from scratch to show that transformers can ICL % ICL as a verb
simple boolean functions, while their performance deteriorates on more complex tasks. 
%\cite{} studied ICL by pretraining a linearly parameterized single-layer linear attention model for linear regression with a Gaussian prior proving that the pretrained model closely matches the Bayes optimal algorithm. 
\cite{raventos2024pretraining} investigated how ICL in models evolves as the number of pretraining examples increases, in a setting where the training and test distributions are identical. % I am really not a fan of this phrasing : "an evenly matched training and test distribution" or something like that is better. 

\cite{olsson:etal:2022} offer an in-depth analysis of ICL across tasks using a general evaluation measure on prompt length. They propose that ICL is enabled by a learned copying and comparison mechanism for ICL. %Uh this sentence is clearly missing a verb, because it doesn't mean anything like that, structurally, it's just "They propose that a mechanism". is ICL enabled by the mechanism ? 
In Section \ref{sec:learn}, we show their proposal does not fit our task.

%Finally, \cite{naim:asher:2025} published an ICL study on continuous functions.  We have adopted their distinction between two types of ICL, ICL$_1$ and ICL$_2$, which captures an important distinction we argue for here.  We do an in depth study of the case of linear functions, whereas \cite{naim:asher:2025} study higher order polynomial and continuous functions.
%\cite{garg:etal:2022} worked out the details on the learning set up for ICL of ${\cal L}$
% \\
%\includegraphics[width=15cm]{training_figure.png} \\
%   They used squared error as the loss function and sampled a batch of random prompts at each training step and update the model through a gradient update. They used a batch size of 64 and train for 500k steps. The training was done from scratch on the model. They trained three sizes of transformers (layers l, attention ah, and embedding  size e): (3, 2, 64); (6,4, 128); and (12,8,256).   {\color{blue} je pense qu'on doit parler moins en detail}% with a restricted embedding size (256 instead of 768).

 %Although 1 attention layer only models do not have this feature (and so by implication do not really ICL according to them), when induction heads are induced in such models, their ICL performance improves.  We have noticed that ICL for ${\cal L}$ occurs even with 1 AH and 1 MLP player.

\section{Experimental setup} %typo: setupv

We investigate transformer models that are trained directly on an in-context learning objective. We trained several small decoder-only %you need an hyphen "decoder-only"
transformer models from scratch to perform ICL of linear functions.\footnote{Our code can be found in https://github.com/omyokun/re-examining-LF-ICL} Our models feature from 1 to 18 layers (L) and from 1 to 8 attention heads (AH) with an embedding size of 64 to 256. 

The model's task is to predict the next value for $f(x_i)$ through a prompt of type $(x_1,f(x_1),...,x_i)$. We refer to that prediction as $\fh(x_i)$.  To train the model to perform ICL % Hum hum... verb
${\cal L}$, we looked for a $\theta^{*}$ that optimizes the following auto-regressive objective: $$ \theta^{*} = \argmin _\theta \mathbb{E}_{x_i \in D_I , f \in D_F} \small{\left[ \sum_{i=0}^k \ell
\left(f\left(x_{i+1}\right), {\cal L}_\theta\left((x_1,f(x_1),\dots,f(x_i), x_{i+1})\right)\right)\right]} $$  

%I just changed the "..." by \dots

where ${\cal L}_\theta$ is a ``learner'', $\ell$ is squared error and $f : x \rightarrow ax + b$ is a linear function with $a,b$ sampled from some training distribution $D_F$ for functions. Samples $x_i$ are sampled from  a training distribution for points $D_I$.  We note that $f \in D_F$ and 
$x \in D_I$. At each training step, we choose at random, a function $f \in D_F$ and then a sequence of points $x_i \in D_I$  from a distribution $D_I$. We use a batch size of 64 and train for 500k steps. The models saw over 1.3 billion training examples for each distribution we studied.  For $D_F$ and $D_I$ we used several distributions: the normal distribution $\mathcal{N}(0,1)$, uniform (``rectangle") distributions over specified intervals, and bimodal distributions. %it will be "rectangular" as it's the adjective and also it sounds a bit like it's 2 different things (I think you mean it's just the two names of this distributions) so maybe "uniform (rectangular) distributions" or "uniform (also know as "rectangular") distribution" but the later cut the flow for a long time  
% also studied by \cite{garg:etal:2022} but also chose Gaussian test distributions that had different parameters.  We also looked at . 

We evaluate the models across various test distributions for both function classes $D^t_F$ and data points $D^t_I$. During training, we always set $D_F=D_I$.  For testing, however, we sometimes consider scenarios where $D^t_F \neq D^t_I$. To evaluate ICL performance of a model on $(D_I^t, D_F^t)$, we generate a set of $N=100$ functions in $D_F^t$. Our data samples for testing are composed of $N_b = 64$ batches, each containing $N_p = 41$ points in $D_I^t$. For each function $f_j$, in each batch b, for all points, we predict for each $x_k^b$, $k \geq 2$, $f_j(x_k^b)$ given the prompt $(x_1^b, f_j(x_1^b),..., x_{k-1}^b , f_j(x_{k-1}^b), x_k^b)$. We then calculate for each function the mean average over all the points $N_p$ of all batches $N_b$, then do a mean average over all functions.  Formally this gives us: $$\epsilon_\sigma = \frac{1}{N} \sum_{j=1}^{N} \sum_{b=1}^{N_b} \frac{1}{N_b} \left(\frac{1}{N_p} \sum_{k=3}^{N_p} (\text{pred}_k^{b,j} - f_j(x^{b}_{k}))^2\right)$$

We also have to define an \textit{error rate} for our problem.  The error rate we chose is: $r_\epsilon = \frac{\epsilon_\sigma}{|\epsilon_* - \epsilon_0|}$ where $\epsilon_*$ is the best $\epsilon_\sigma$ error for a model M with $\fh(x)$ calculated with Least Squares, and $\epsilon_0$ is the worst $\epsilon_\sigma$ error for a model $M$ such that $\fh_M(x) = 0$, $\forall x$.  In all error calculations, we exclude the first two predictions of each batch from the squared error calculation.  We need at least two points to be able to find a linear function and the first two predictions by the model are hence almost always wrong. %A drawback of this method is that if a batch gives abnormal predictions, its error will be diluted by all the other batches and we won't notice it. %So we also calculated an average error $\delta$ over batches.  

\section{Models ICL$_1$ but do not ICL$_2$ ${\cal L}$}\label{sec:icl}

When trained on $D_F = D_I = \mathcal{N}(0,1)$ and tested on $D^t_F = D^t_I = \mathcal{N}(0,1)$, even many small models were able to converge to near-zero average error.  %weird phrasing, I would say "to an average close to zero" (or 0 if you want to keep digits as number)
Error rates fluctuated across batches and were not consistently zero, %I get what you mean but you struggled to phrase it simply: "error rate varied between batches and was not always zero" (or keep 0)
and so similar to \cite{liu:etal:2024}'s findings on MSE estimation by transformers.

\begin{figure}[!h]
\center 
\includegraphics[width=8cm, alt={This is a plot showing the evolution of error rates for different models. All models are trained on data with input and output distributions drawn from a standard normal distribution, while the test target data follows a normal distribution with varying standard deviation.}]{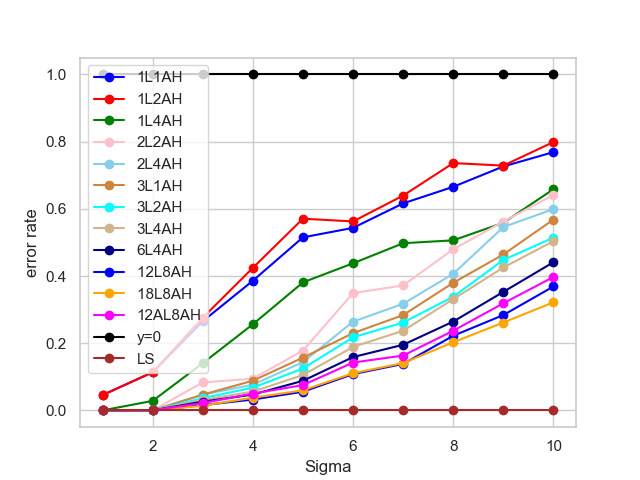}

\caption{Evolution of error rates for various models  with $D_F = D_I = D^t_I = \mathcal{N}(0,1)$ and $D^t_F$ for various $\mathcal{N}(0, \sigma)$. $y=0$ illustrates a model that predicts $f(x_n) = 0, \forall f$ and $\forall x_n$. And LS represents linear or ridge regression, which is trivially a perfect estimator given our totally clean input data.  
\label{progressive-loss}}
\end{figure}

These results hold for full transformer architectures models, as well as for models without feedforward networks.  In fact the 12-layers, 8-attention heads model without feedforward networks performed equivalently to the full transformer model (see Figure \ref{progressive-loss}).  Together these were our best models. However, models with only MLP layers, were unable to do the task. %\footnote{It may be interesting to look at the results for attention layers when they are provably causally effective, as with efficient attention \cite{naim:asher:2024a}.}  
%{\color{blue} est-ce qu'on retire cette observation? pour eviter de se repeter avec l'autre papier}
%{\color{orange}
%\begin{obs} \label{obs:restricted} transformer models can in principle ICL$_1$
%any linear function f over a small interval [a, b] if we pick a suitable training and testing regime. 
%\end{obs}
%}
\begin{obs} \label{obs:attention}
Attention layers are necessary and sufficient to ICL$_1$ the class of linear functions. Models with only MLP were unable to ICL$_1$.
\end{obs}

% but ICL only worked reasonably well when the model had 2 AL (see also figure \ref{big30}).  
% Attention-only models could ICL linear functions reasonably well, at least in when $D_F = D_I = D^t_F = D_I^t$.  For more details see the discussion in Appendix D. 

Varying the test distributions, however, altered performance considerably.  All the models had systematic and non 0 average error once we chose $f \in D^t_F = \mathcal{N}(0, \sigma$) for $\sigma > 2$.  While \cite{zhang:etal:2024} claim shifting the distribution sampled at inference of the functions does not affect performance, Figure \ref{progressive-loss} clearly shows that for transformer models with soft attention, this task shift at inference reduces performance dramatically.  Figure \ref{progressive-loss} shows that the error rate increases substantially and non-linearly as $D_F^t = \mathcal{N}(0,\sigma)$ and $\sigma$ increases.  To ensure that comparisons between models are meaningful, for each $\mathcal{N}(0,\sigma)$, we set a seed when generating the 100 random linear functions, ensuring that each model sees the same randomly chosen functions and the same set of prompting points $x_i$.    
%Figure \ref{progressive-loss} plots the evolution of error loss for various models.
The Table \ref{table:3} in the Appendix contains the full scores for average error. 

 %such a shift affects the results in an important way, where we take $\mathcal{N}(0,1)= D_F$ ( but $D^{t}_F = \mathcal{N}(0, \sigma)$ for $1 \leq \sigma \leq 10$. %The full error scores are in the Appendix.

\begin{figure}[!h]
%\includegraphics[width=11cm]{figures/evolution of errorsbis.png} 
%\center 
%\includegraphics[width=8cm]{figures/errors22.png}
%\includegraphics[width=8cm]{figures/hmaps.png}
\includegraphics[width=4cm, alt={Heatmap showing log error rates for a model trained with both input and weight data drawn from a normal distribution with mean zero and standard deviation one. During testing, inputs and weights are sampled independently with varying standard deviations from one to ten. Darker shades indicate lower error rates, and the shading scale is adjusted for interpretability.}]{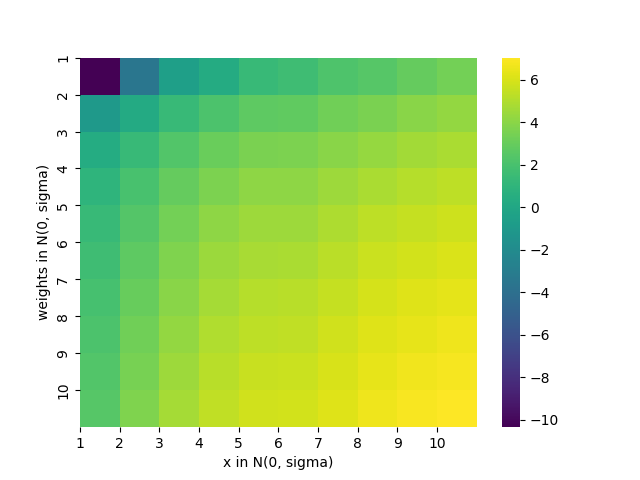}
\includegraphics[width=4cm, alt={Heatmap showing log error rates for a model trained with both input and weight data drawn from a normal distribution with mean zero and standard deviation ten. During testing, inputs and weights are sampled independently with varying standard deviations from one to ten. Darker shades represent lower errors, and the color scale is adapted to enhance comparison.} ]{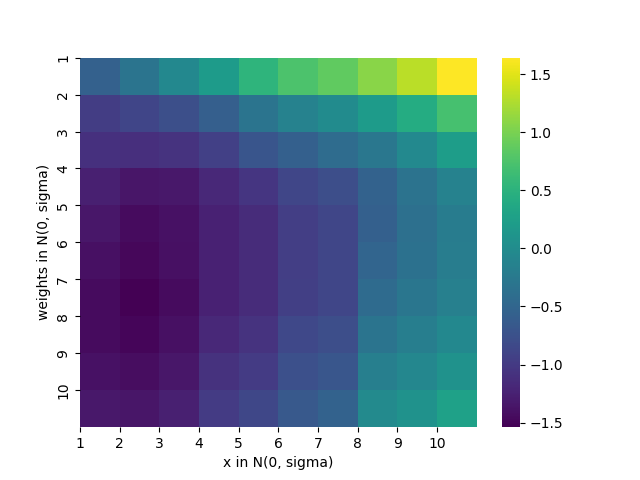}
\includegraphics[width=4cm, alt={Heatmap showing log error rates for a model trained with both input and weight data drawn from a normal distribution with mean zero and standard deviation one hundred. During testing, inputs and weights are sampled independently with varying standard deviations from one to ten. Darker shades indicate better performance, but note that shading values differ from the other plots for better clarity.}]{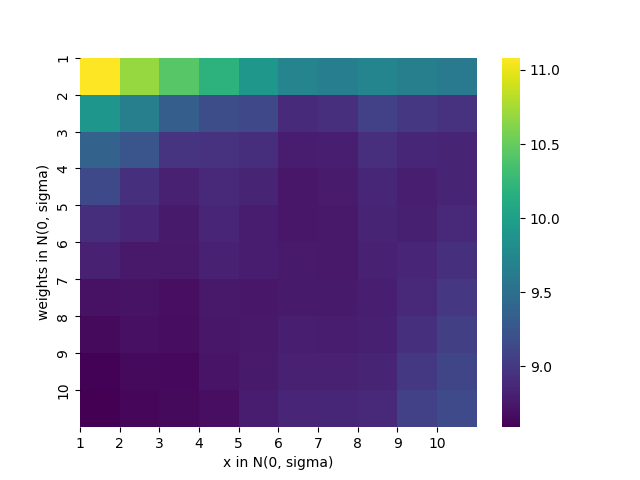}
\caption{Heatmaps showing the evolution of log error rates for a model trained on $D_I=D_F=\mathcal{N}(0,1)$ (left), one trained on $D_I=D_F=\mathcal{N}(0,10)$ (middle) and one trained on $D_I=D_F=\mathcal{N}(0,100)$ (right). During testing, inputs and weights are drawn independently from $D^t_I \sim \mathcal{N}(0,\sigma_1)$ and $D^t_F \sim \mathcal{N}(0,\sigma_2)$, with $\sigma_1, \sigma_2 \in \{1, \dots, 10\}$.  Note that while darker shades are always better, the shades have different values in the figures above to help with interpretability.
\label{hmap}}
\end{figure}

The heatmaps in Figure \ref{hmap} show how our models generalized outside of three different sets of training distributions. Figure \ref{hmap} shows that the generalization ability of models trained on $D_F = D_I = \mathcal{N}(0,1)$ rapidly degrades once we sample elements far away from the training distribution, such as when we set $D^t_F$ and $D^t_I$ to $\mathcal{N}(0, \sigma)$ for $\sigma > 1$.  
On the other hand, while the errors in the performance of models trained on $D_F = D_I = \mathcal{N}(0,100)$ are large when tested on $\mathcal{N}(0,1)$, performance improves as we move toward testing regimes that take elements $x_i, f(x_i)$ within intervals [a,b] that contain a higher proportion of examples seen in training (Figure \ref{hmap}).  This trend is even clearer in the heatmap for the model trained on $\mathcal{N}(0,10)$ in the same figure.  The model shows relatively good performance on $D^t_F = D^t_I = \mathcal{N}(0,1)$ though not as good as the model trained on $\mathcal{N}(0, 1)$.  However, the model trained on $\mathcal{N}(0,10)$'s performance  improves as the testing regime comes to include a higher proportion of functions seen in training.  During testing, we also observed that models trained on distributions with a larger variance\footnote{A larger variance means that a larger interval includes a large majority if not the totality of the examples seen in training.} provided better generalization performance in the sense that such models had lower squared errors than models trained on distributions with a smaller variance.  Thus, a model trained on $\mathcal{N}(0,10)$ has a better generalization ability and more robust performance than the model trained on $\mathcal{N}(0,1)$.

\begin{table*}[!ht]
\small{
\begin{tabular}{l|l|l|l|l|l|l|l|l|l|l}
 \hline
  models \ $\backslash$ \ $\sigma$ & 1 & 2 & 3 & 4 & 5 & 6 & 7 & 8 & 9 & 10 \\ 
 \hline\hline
 $1L1AH_N$ & 0.1 & 0.8 & 5.1 & 13.1 & 26.9 & 39.7 & 53.0 & 84.8 & 120.0 & 153.2 \\

 %$1L2AH_$   & 0.1 & 0.8 & 5.3 & 14.4 & 29.8 & 41.1 & 55.0 & 93.8 & 120.4 & 159.2 \\

 $1L4AH_N$   & 0.0 & 0.2 & 2.7 & 8.7 & 19.9 & 32.0 & 42.8 & 64.5 & 92.3 & 131.2 \\
 \hline
 $2L1AH_N$  & 0.0 & 0.1 & 2.0 & 4.9 & 13.7 & 27.0 & 36.1 & 64.9 & 99.0 & 134.0 \\

% $2L2AH_N$  & 0.0 & 0.0 & 1.6 & 3.2 & 9.3 & 25.5 & 32.0 & 61.1 & 92.9 & 127.8 \\

 $2L4AH_N$   & 0.0 & 0.0 & 0.9 & 2.6 & 7.5 & 19.3 & 27.3 & 51.8 & 90.2 & 119.4 \\
 \hline
 $3L1AH_N$   & 0.0 & 0.0 & 0.9 & 3.0 & 8.2 & 16.8 & 24.4 & 48.4 & 76.7 & 113.2 \\

% $3L2AH_N$  & 0.0 & 0.0 & 0.7 & 2.3 & 6.5 & 15.9 & 22.5 & 43.1 & 74.0 & 102.5 \\

 $3L4AH_N$   & 0.0 & 0.0 & 0.6 & 1.9 & 5.5 & 13.8 & 20.4 & 42.2 & 70.3 & 100.4 \\
 \hline
 $6L4AH_N$   & 0.0 & 0.0 & 0.5 & 1.6 & 4.6 & 11.6 & 16.8 & 33.7 & 58.3 & 87.9 \\
 \hline
$12L8AH_N$  & 0.0 & 0.0 & 0.3 & 1.1 & 2.9 & 7.9 & 11.9 & 28.3 & 46.9 & 73.5 \\ [1ex] 
 \hline
 $18L8AH_N$   & 0.0 & 0.0& 0.2& 1.1& 2.8& 7.1& 10.3& 22.9& 40.3& 64.6 \\ [1ex] 
 \hline \hline
   $6L4AH_B$,    & 0.01 & 0.04 & 0.23 & 0.44 & 1.19 & 2.15 & 3.08 & 4.8 & 9.98 & 18.01 \\

   $6L4AH_U$,    & 0.02 & 0.04 & 0.11 & 0.24 & 0.57 & 1.36 & 1.82 & 4.62 & 10.23 & 15.07 \\
 \hline\hline
 $12L8AH_N$,   & 0.0 & 0.0 & 0.32 & 1.34 & 3.14 & 8.8 & 12.13 & 30.14 & 49.37 & 73.93 \\  

  \textbf{sorted $12L8AH_N$}  & 0.0 & 0.01 & 0.32 & 1.63 & 3.69 & 8.39 & 10.06 & 27.11 & 43.23 & 58.56 \\  

  \hline
 $12L8AH_B$   & 0.0 & 0.01 & 0.08 & 0.29 & 0.78 & 2.23 & 3.66 & 9.04 & 18.68 & 30.23 \\ 

 \textbf{sorted $12L8AH_B$}  & 0.01 & 0.03 & 0.18 & 0.25 & 0.74 & 2.27 & 2.62 & 6.87 & 13.73 & 20.8 \\ 

  \hline
 $12L8AH_U$ & 0.0 & 0.01 & 0.13 & 0.71 & 1.92 & 6.78 & 10.92 & 27.91 & 38.75 & 64.39 \\ 

 \textbf{sorted $12L8AH_U$}   & 0.01 & 0.01 & 0.13 & 0.75 & 2.12 & 6.18 & 10.5 & 26.8 & 36.3 & 53.48 \\ [1ex] 
 \hline
 \textbf{REF$_{N}$: y=0}   & 2.19 & 7.05 & 19.22 & 33.94 & 52.23 & 73.08 & 86.02 & 127.43 & 165.27 & 199.31 \\ 
  \textbf{$REF_{U}$: y=0}   & 1.52 & 4.43 & 13.55 & 19.94 & 30.81 & 44.75 & 52.71 & 76.11 & 105.43 & 128.52\\
  \hline
  \hline
 3NN  & 0.03 & 0.14 & 0.27 & 0.66 & 1.09 & 1.32 & 1.75 & 2.45 & 2.95 & 4.01 \\ [1ex] 
 \hline
\end{tabular}
}

\caption{Comparison to show the evolution of squared $\epsilon$ type error depending on the distribution according to which we take the parameters, without taking into account the error of the prediction of the first and second prompts. Embedding size for all models is 64 except for 12L and 18L models (256K).   For models N, $D_F = \mathcal{N}(0,1)$; $D_i = D_i^t = \mathcal{N}(0,1)$. For B, $D_F = 0.5{\cal N}(-1,1) + 0.5 {\cal N}(1,1)$ and U, $D_F = {\cal U}(-5,5)$. For B and U testing, $D^t_i = {\cal U}(-1,1)$ and  $D^t_F=\mathcal{N}(0,\sigma)$.  We show error rates for models prompted without and with the natural ordering on the prompts [sorted], for the large model size. 3NN refers to \cite{olsson:etal:2022}'s method which generates values through the average of the 3 nearest neighbors.}
\label{table:3}
\end{table*}
To further test our hypothesis we evaluated the performance of our models using bimodal and uniform distributions for training data. The bimodal distribution $0.5\mathcal{N}(-1,1) + 0.5\mathcal{N}(1,1)$ expanded the range of values $f(x)$ the model will likely see during training.  Models trained on this distribution for $D_F$ or on uniform distributions with a larger span like ${\cal U}(-5,5)$ had more robust performance than models trained with $D_F = \mathcal{N}(0,1)$ at least with $D^t_F = D^t_I = \mathcal{N}(0,\sigma)$ and $n \geq 6$.  The best models had %almost equally good performance on $D^t_F = \mathcal{N}(0,\sigma)$ for $\sigma \leq 3$ and 
superior performance with $D^t_F = \mathcal{N}(0,\sigma)$ for $\sigma \geq 3$ (see  Table \ref{table:3}, in which  $D_I^t = \mathcal{N}(0,1)$).  But they also had limited generalization capability. For error values on various distributions and models see Table \ref{table:3}.% in Appendix \ref{sec:appendixB}.

To sum up our findings, 
\begin{obs} \label{obs:density} (i) Models have better performance over intervals that contain a larger proportion of examples in the training distribution.  (ii) Models trained on a distribution with larger variance (up to a certain point) had better generalization ability but less accuracy than models trained on distributions with smaller variance.
\end{obs}

We also tested an alternative hypothesis to that in Observation \ref{obs:density} on which the total number of points observed in the distribution, rather than the proportion, is the critical factor. We trained one model on ${\cal U}(-1,1)$ for $100k$ steps and another on ${\cal U}(-5,5)$ for $500k$ steps, ensuring that both models were exposed to a statistically equivalent number of points within the interval $[-1,1]$. The results remained similar, showing that the total number of points observed is not the critical factor.\footnote{A checkpoint analysis further confirmed this indicating that optimal performance is typically reached at a fixed training step number, beyond which model performance fluctuates and does not really improve.}

Model performance on out-of-distribution examples also improved somewhat when the sequence of $x_i$ in the training prompts ($p_1$) was sorted, to follow the natural order on $\mathbb{R}$, especially for larger models.  Error rates with sorting are lower, by up to a third—compared to error rates without sorting for small values of $\sigma$, with $D^t_F = \mathcal{N}(0, \sigma)$. We also tested prompts ($p2$) on a  12L8AH model in a kind of "curriculum learning" ; models with $p2 < p1$ did less well than models with $p1 = p2$.  

\begin{figure}[!h]
\center 
\includegraphics[width=7cm, alt={This figure shows a hateamap of the evolution of attention weights in the last layer of a 12L8AH model}]{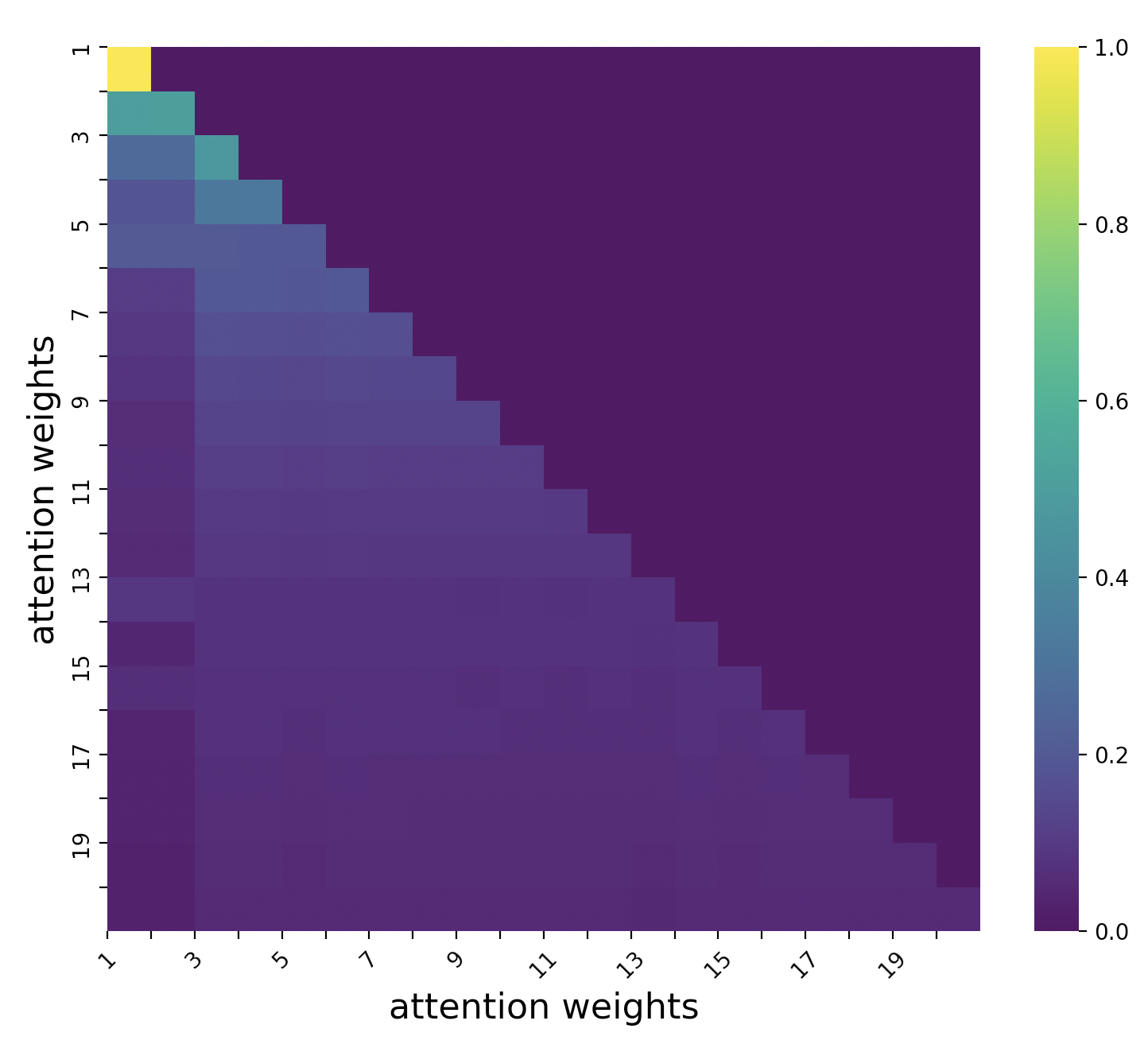}
\caption{Heatmap showing the evolution of attention weights in an attention head of the last layer of 12AL8AH model
\label{hmapattnweights}}
\end{figure}

We did a heatmap analysis of the weights of the elements in the attention matrices of the last layer of our model.  The results are in Figure \ref{hmapattnweights}.  
As can be seen, the model learns to pretty much ignore the first element and its prediction on that (the first two elements in its context) after the 6th prediction. Attention weights are largely uniform over the rest of the elements in the sequence.  This suggests that it uses most if not all contextually given pairs of input function values, $x_i, y_i = f(x_i)$, to predict a value $y_n$ for input $x_n$.

All this leads to the following observation. 
\begin{obs} \label{obs:sequence} A model takes into account the whole sequence to predict $\fh(x_n)$, not just some small, fixed subsequence.%, and uses the points to update its estimates of $\fh$. 
\end{obs}

Contrary to what \cite{akyurek:etal:2022,vonoswald:etal:2023} have suggested, our observations show that our models did not use linear regression to ICL the target function. If they had, we would not observe the error patterns or variations in performance, nor would the models disregard the class form of the target function in making their predictions; otherwise, the use of all elements in the sequence would have been superfluous.  In conclusion, all models tested failed  to ICL$_2$ ${\cal L}$.  Moreover, we have seen a trade off between accuracy and generalizability from Observation \ref{obs:density}; given current transformer models, we cannot have both accuracy and generalizability.

\section{Error analysis and boundary values}\label{sec:4.4}
\begin{figure*}[!h] 
\includegraphics[width=4cm, alt={Plot showing model 12L8AH trained on input and weight data from a standard normal distribution, evaluated on the function f of x equals x applied to high input values. The plot illustrates how the model performs when the linear relationship holds primarily in the upper range of x.}]{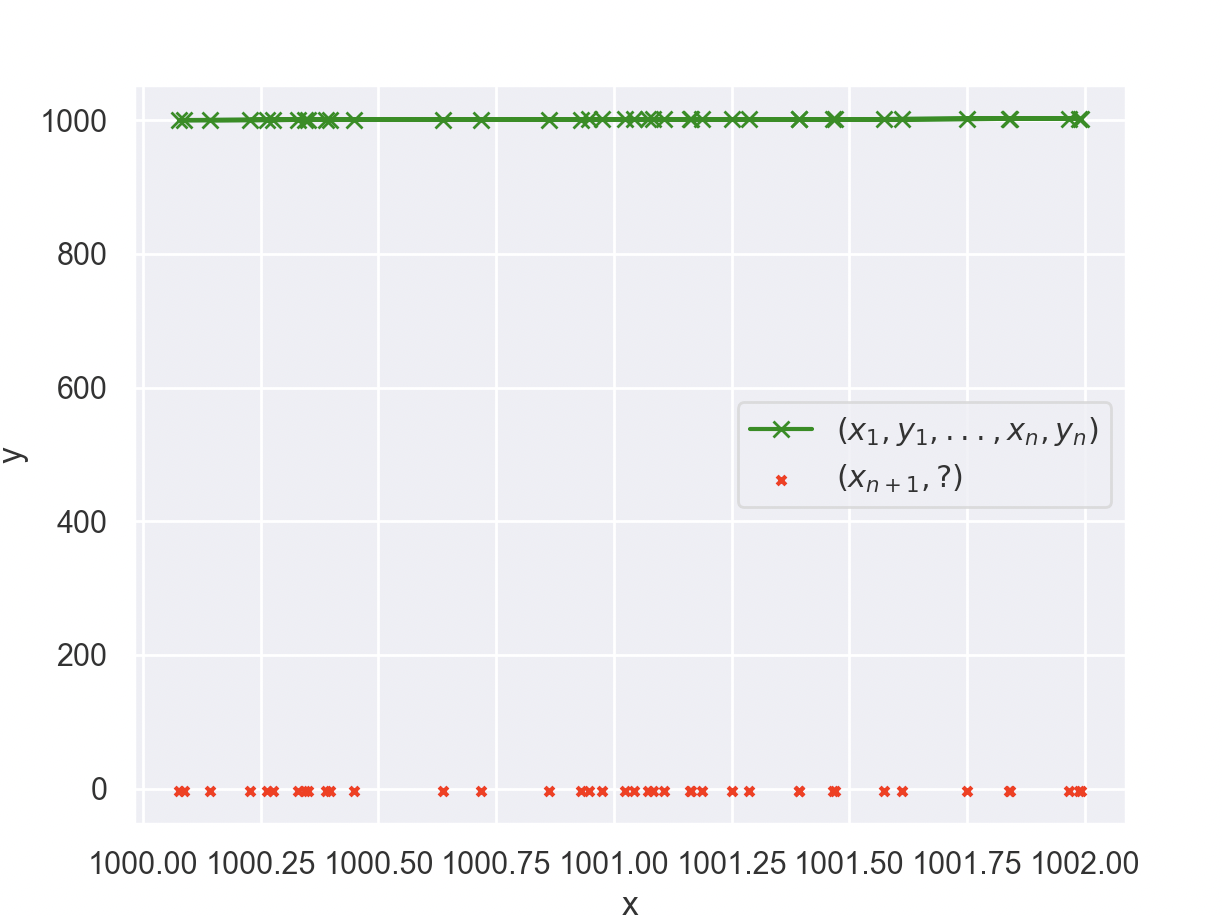}
\includegraphics[width=4cm, alt={Plot showing model 12L8AH trained on standard normal input and weight data, evaluated on the function f of x equals ten times x in the typical or mid-range of input values. This plot focuses on how the model handles a steeper linear relationship across the central portion of the input domain.}]{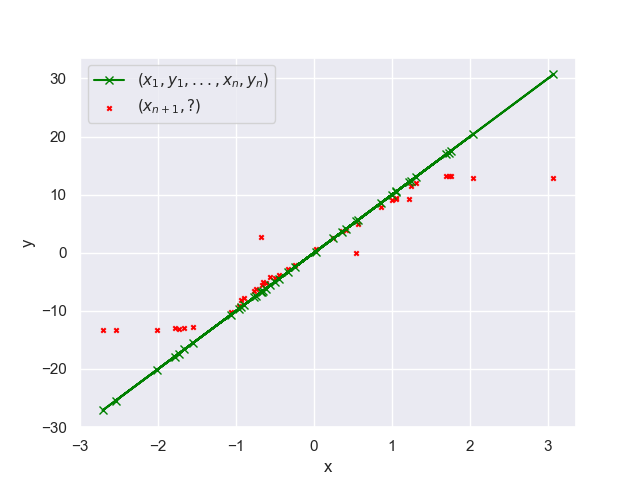} 
\includegraphics[width=4cm, alt={Plot showing model 12L8AH trained on input and weight data from a standard normal distribution, evaluated on the function f of x equals ten x applied to low input values.}]{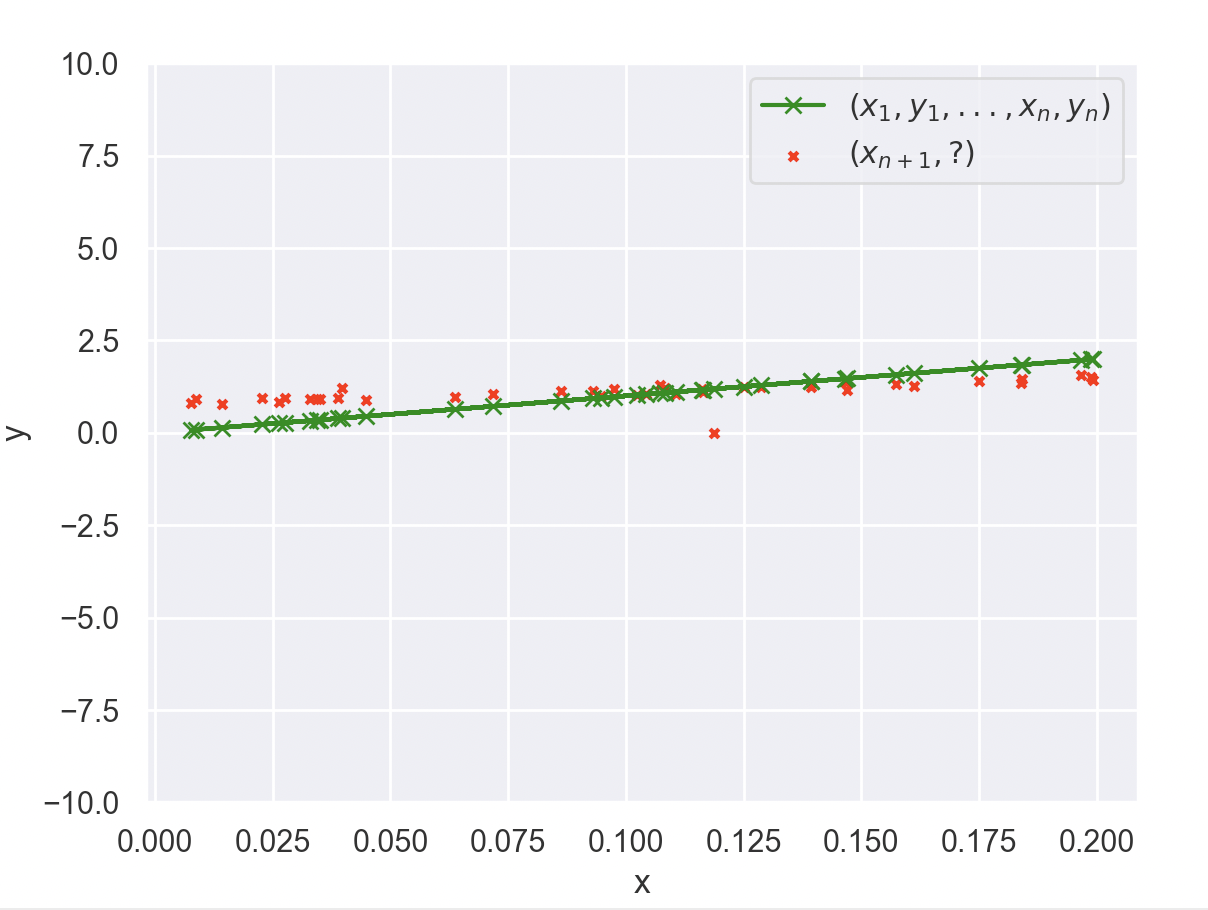}
\caption{Plots for model  12L8AH, trained on $D_I = D_F = \mathcal{N}(0,1)$  for $f(x)=x$ for high values (left) of $x$ and $f(x)=10x$ for normal (middle) then for low values of $x$ (right)}\label{sequence}
\end{figure*}

During all our experiments, we noticed that all our models presented what we called \textit{boundary values}, which are values that the model cannot exceed during inference predictions (see Figure \ref{sequence}). These \textit{boundary values} act as a filtering mechanism, restricting the model to generate outputs only within a specific range. This constraint prevents the generation of values beyond these limits, effectively preventing the model from generalizing its good performance on the task over a small interval to larger values outside that interval. This behavior can be seen also in \cite{giannou:etal:2024}'s figures, despite their use of a transformer based on linear attention. 

Since these models are trained from scratch, the only factor that can determine the boundary values is the training process itself. We conjecture that the model memorizes some bounds that are the largest and smallest values encountered during training, and then sets them as boundary limits.  This effectively functions as a bandpass filter. 

\begin{figure}[ht!]  
\centering
\includegraphics[width=0.4\textwidth, alt={Plot showing the performance of the 12L8AH model trained on input data sampled uniformly between -5 and 5, when evaluated on the function f of x equals 9 times x.}]{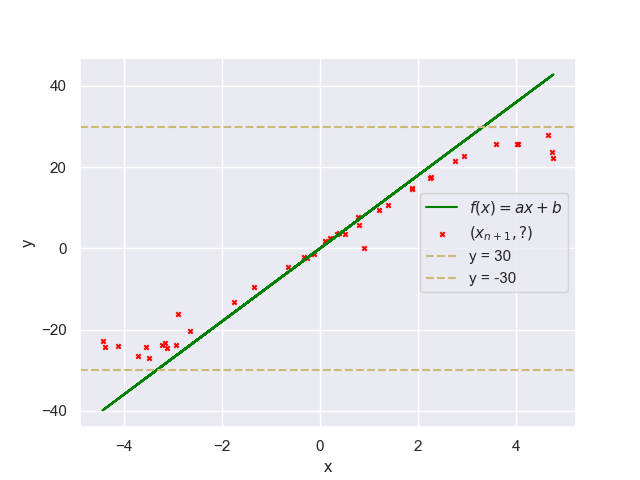}
\includegraphics[width=0.4\textwidth, alt={Plot showing the performance of the 12L8AH model trained on input data sampled uniformly between -5 and 5, when evaluated on the function f of x equals 40 times x plus 40.}]{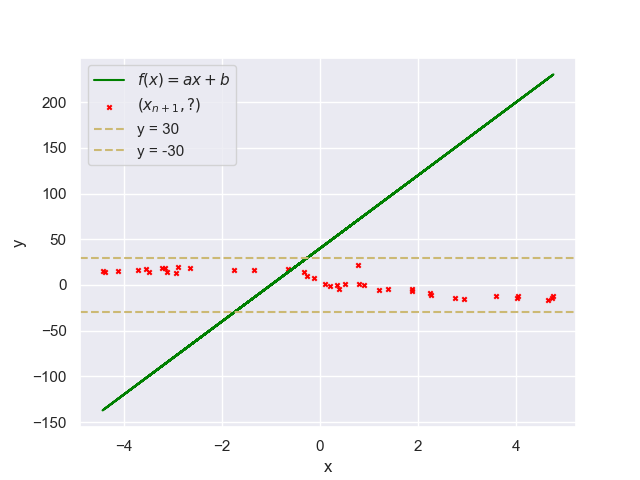}
\caption{Plots for $f(x) = 9x$ and $f(x) = 40x + 40$ for a 12L8ah model trained on ${\cal U}(-5,5)$ \label{40x+40}}
\end{figure}

However, determining the smallest and largest values a model could have encountered by training for data and coefficients in Gaussian distributions is impossible.  As a result, to have control over this parameter, we trained from scratch models on uniform distributions.  These experiments enabled us to confirm our hypothesis that boundary values correspond to the smallest and largest values that the model has encountered and memorized.  One advantage of training on a uniform distribution is that we know exactly what the smallest and largest values that the model could have seen in its training. For example, setting $D_F, D_I$ to ${\cal U}(-5,5)$, the largest value the model could have seen is $30 = 5\times 5 + 5$ %use "\times" instead of *
and the smallest value it could have seen is $-30$.  Most likely it saw values significantly $> -30$ and $< 30$.  All our models trained on ${\cal U}(-5,5)$ estimate the target function more or less well for $x$ with $f(x) \in [-30, 30]$.  But once we took
inputs $x_i$ or coefficients of our target function $f$ to force $f(x_i) \not\in [B^{-}, B^{+}]$, the estimations become either constant functions or chaotic. Figure \ref{40x+40} with equation $f(x) = 40x + 40$ provides another example. 

 Boundary values hold for all transformer models tested including attention only models (for plots see Figures 6 and 7 in the Appendix).
 %\ref{big30} and \ref{fig:boundary}
Linear transformer architectures also show boundary values  \cite{giannou:etal:2024}.  For example with $D_F = D_I = {\cal U}(-5,5)$, consider as an illustrative example the target function, $f(x) = 9x$, with our largest trained model. The model approximates $f(x)$ well within a certain range $[B^{-}, B^{+}] = [-30,30]$ that corresponds to the maximal and minimal values it could have seen.  On the other hand, it predicts $\fh(x)$ as 
 a constant function for $x$ such that $\fh(x) \not\in [B^{-},B^{+}]$
 within a certain range (See Figure \ref{40x+40}).  
 Indeed, model errors depended on how many functions and values sampled forced the result outside $[B^{-}, B^{+}]$, as well as on the proportion of a model's training sequences in the test distribution (see Observation \ref{obs:density}). %We call values $B^{-}, B^{+}$ %like -30 and 30 where the model starts to predict constant functions $\fh(v)$ and $\fh^-(v)$ {\em boundary values}.
% By training on uniform distributions, we can determine the boundary values exactly; e.g, for $U(-5,5)$ $B^{+} = 5\times 5 + 5$ and $B^{-} = -5\times 5 - 5$.  These are the biggest and smallest values the model could have seen during training. If such a model hasn't seen a value above $B^{+}$ or below $B^{-}$, it won't infer one.  Different models trained on different uniform distributions give different boundary values (see below). \\
%The plots show that the model's prediction $\fh(x)$ diverges dramatically from $f(x)$ outside of a certain interval, but it  within that interval the prediction is good.  For functions $f(x)$  sampled far outside $\mathcal{N}(0,1)$ (for example $f(x) = 30x + 30$ and $D^t_I = \mathcal{N}(0,1)$, however, the results are catastrophic and similar to those in the first plot of Figure \ref{sequence}. (See Figure \ref{big30} in the Appendix).

As shown in the left plot of Figure \ref{40x+40}, $\fh^+(v) \approx 30$ for values $v$ such that the ground truth target function $f$ satisfies $30 \leq f(v)$, and the model predicts an approximately constant function $\fh^-(v) \approx -30$ for values $v$ on which $f(v) \leq -30$.  Different models trained on different uniform distributions give different boundary values (see below).  
To summarize:
\begin{obs} \label{obs:boundary} (i) Boundary values $B^-,B+$ are determined by maximal and minimal elements seen during training.   (ii) For all models $M$ and for values $v$ such that $ B^{+} < f(v) < B^{+} +\alpha$, where $\alpha$ is a constant determined by $M$, $\fh_M(v) \approx B^{+}$, and  for $ B^{-} - \alpha < f(v) < B^{-} , \fh_M(v) \approx B^{-}$.  (iii) For functions and data samples when the values of $f(v)$ in the prompt sequence are such that $f(v) > B^{+}+\alpha$ or $< B^{-} - \alpha$, the model assigns $\fh(v)$ a random value within $[B^-, B^+]$.
\end{obs}
The parameter $\alpha$ in Observation \ref{obs:boundary} accounts for the fact that larger models trained on the same distribution and the same number of data will ICL ${\cal L}$ functions over a slightly larger number of intermediate values than smaller models, as Figure \ref{progressive-loss} suggests. Figure 7
%\ref{fig:boundary}
in the Appendix shows plots of
the predictions of two models (12L8AH, and 6L4AH) for $D_F, D_I = \mathcal{N}(0,1)$ for target $f(x) = 10x$.  The larger model has boundary values $\approx$ $-13.7, 13.7$, the smaller one boundary values $\approx -12, 12$.

Since a model's predictions do not go beyond the boundary values defined by training, a consequence of Observation \ref{obs:boundary} is that error rates depend on the distance of the target function’s values from the boundary values and hence the majority of the data points in the model’s training. Figures \ref{progressive-loss} and \ref{hmap} verify this progressive increase in error rate as the testing distribution gets larger and larger.  

Training with distributions over much larger intervals, for example $\mathcal{N}(0, 10)$ or $\mathcal{N}(0,100)$, can extend boundary values.  Models trained on ${\cal N}(0, 10)$ show much better generalization ability than the models trained on $\mathcal{N}(0,1)$.  On the other hand, they are less accurate
on small values as Figure \ref{hmap} shows.  Models trained with $D_F = {\cal U}(-10,10) = D_I$ generalized even better, but their performance on $\mathcal{N}(0,1)$ was significantly worse
than models trained on $\mathcal{N}(0,1)$.  The $\mathcal{N}(0,100)$ performed poorly in our testing scenarios but improved as values got larger. Our observations thus present a dilemma: training on restricted intervals is needed for high accuracy, but it inevitably hampers generalizability.  Generalizability requires training over larger intervals but that reduces accuracy.
 %, though performance deteriorated substantially otherwise.  

As evident from Section 2, we are not the first to investigate out of distribution behavior.  \cite{giannou:etal:2024} investigated out of distribution behavior but only on $D_I \neq D^t_I$ (not shifts in $D_F$).  Unlike \cite{giannou:etal:2024}, we found that larger models had better performance than smaller ones, when $f(x), x$ were within boundary values.  Our results also differed from \cite{zhang:etal:2024}'s shifting from $D_I$. They shift the prompt distribution but not that of the query.  We shift both prompt and query distributions. With our shifts, we found that the choice of points is important and model performance degrades considerably when the values of the functions on the chosen points lie beyond the boundary values, which \cite{zhang:etal:2024} do not.  As far as we know, we are the first to take boundary values and their dependence on model parameters as key indications of what is actually going on in ICL.
\section{What and how are the models learning?}\label{sec:learn}

In Section \ref{sec:icl}, we gave our reasons why the hypotheses of \cite{akyurek:etal:2022,vonoswald:etal:2023} about models being able to ICL linear functions via linear regression are incorrect.  %ed us to expect that a transformer model given $(x_1,f(x_1),...,x_n), ?)$ would perform a linear regression to ICL a linear function. 
The boundary values $B^-, B^+$ of Section \ref{sec:4.4} provide further evidence that models are not performing linear regression to define a linear function; otherwise, the models' predictions within $[B^-, B^+]$ should carry over to intervals larger than those defined by the majority of their training samples.  A lack
of full generalizability is perhaps expected\footnote{A linear function is a $\Pi^0_1$ set in the Borel hierarchy, which \cite{asher:etal:2023} show is not learnable using ordinary LLM assumptions.} but the limitations on learnability we have seen are quite strong.   

The presence of boundary values suggests that our models have memorized at least a portion of their training.  Do our models simply overfit the data?  We argue no. The pretraining data has no noise, and it is too large to be memorized by our models (our largest models with 256 size embeddings have $32$M parameters; each parameter would have to encode on average sequences for over 100 different functions).  Moreover, our models performed similarly on several different training distributions for $D_F$ and $D_I$ and tested on ${\cal N}(0,\sigma)$ for $\sigma \in \{1,2\}$.  To test overfitting, we added noise to our input, but that did not improve error rates. This implies that the models did not overfit to their training regimes.  Our models seem to reason as well. Given 100 samplings with $D^t_F = {\cal N}(0,1)$ nets on average 20 functions with coefficients the model with $D_F = D_I = {\cal U}(-1,1)$ has not seen in training, we would expect the model's performance to degrade more substantially than it did if only memory were involved. Moreover, models trained on ${\cal N}(0,10)$ had substantially improved generalization ability over ${\cal N}(0,\sigma)$ for all $\sigma$ we tested, as can be seen in Figure \ref{hmap}, even though the odds of having a sequence from training repeated on the test are very low.

%\section{How might the models be learning?}

%Our experiments have isolated several factors about ICL in transformer models: (i) their predictions for a sequence $x_1, f(x_1), x_2, f(x_2), ... x_n, ?$ are sensitive to the number of times it has seen similar sequences $y_i_1, f(y_i_1), y_i_2, f(y_i_2), ... y_i_n, ?$ in its training data.  Our models have certain boundary values $-B, B$ beyond which they have difficulty approximating $f(x)$ for $f(x) < -B$ or $f(x) > B$,
  To go back to the question of what models are learning, \cite{olsson:etal:2022} conjectures that an attention layer only model generates values by averaging the three nearest neighbors. We % need a capîtal letter
  tested \cite{olausson:etal:2023}'s conjecture by using it to calculate the output.  The results were too far from the values we got experimentally (Table \ref{table:3}).  Using \cite{olsson:etal:2022}'s suggestion also gives results that make the wrong predictions for points at or around \textit{boundary values} and other points (see Figure
  \ref{sequence} of $f(x)=x$ for high values in the Appendix).

Instead given a prompt sequence $ (x_{1}, y_1 (=f(x_1)), x_{2}, y_2, \cdots,x_{n}, ?)$, we hypothesize that the model optimizes a projection $\pi$ based on stored projections $\pi_n: Z \mapsto Y$, where $Z$ are inputs it has seen and $Y$ are the corresponding outputs $y_i$ it has seen paired with $x_i$ in $X$
, to predict a value for $y_n \in [B^-, B+]$.  Given Observation \ref{obs:attention}, this projection can be fully determined by the attention layers in the model. 
\hidden{
 To go into a few more details,  let $\vec{x}^l=(x_1^l,...,x_n^l)$ be the input that goes through the multi-head attention at layer $l$. The output \cite{naim:asher:2024a} of multi-head attention after going through k attention heads, then a Linear layer is : $(C_1^l, C_2^l,...,C_n^l)$ where \begin{equation} \label{attn} C_i^l  = \sum^k_{h=1} \sum^n_{j=1}W_O^{l} softmax(\frac{(W_Q^{h,l}x_i^l)^{T}(W_K^{h,l}x_j^l)}{\sqrt{d_k}})(W^{h,l}_Vx_j^l) 
 \end{equation}

\noindent 
where $W_Q^{h,l}, W_K^{h,l}, W_V^{h,l}$ are Query, Key and Value weights matrices respectively in attention head $h$ in layer $l$, $d_k$ is the dimension of the key matrix and $W_O^{l}$ is the matrix of the linear layer that comes after the attention heads in the layer l.  All those matrices have fixed values from training.  The output of the attention layer is a distribution over possible values for $\pi(x_n)$.
}
Our observations show that more attention heads, but even more importantly more attention layers, improve model performance.  
This is, we conjecture, because they provide more memory for storing nearby sequences from which the model can draw to optimize $\pi$ . To sketch a possible projection, we mimic the autoregressive reasoning.  We start just with a target input $x^t_1$.  The attention matrices have encoded a projection in their weights that takes $g_0: x_1^t \mapsto z^m_i$, where $z^m_i = x^t_1 \pm e$ for which there is also a stored projection $p_0: x^m_i\mapsto y^m_i$. It then sets $\pi_0$ such that $\pi_0(x^t_1) = p_0(g_0(x^m_i)) = y^m_i \pm d$, where $d$ is close to $e$.  This is usually wrong, since there are many possible choices for $g_0$ and $p_0$ from training.  Now suppose the model has received $x^t_1, y^t_1, x^t_2$. It now tries to predict $\pi_1(x^t_2)$.  To do this it must find projections $g_1$, $p_1$, and $\pi_1$ such that $\pi_1(x^t_2) = p_1 \circ g_1(x^t_2)$ and $\pi_1(x^t_1) = y^t_1$.  Several choices for $p_1, g_1, \pi_1$ may obey these constraints and so the model may interpolate between them.  $\pi_1$ at level $l$ is defined via the context matrix given by attention \cite{naim:asher:2024a}.
%weights in the sum that gives $C^l_i$ in Equation \ref{attn}.   How accurate the projection $\pi$  will depend, given how we have characterized $\pi_1$ in terms of $p_1,g_1$, on the proportion of training examples in the test distribution.  A higher proportion increases the probability that the model will find a close element of $x^t_2$ using $g$ and a relevant $p_j$.  As the auto regressive inference continues to choose new $\pi$ for $x^t_i$, the presence of more constraints should provide a better approximation $\pi$ of the target function.  In this way the model uses many if not all but the first two elements of its context to determine the projection.  However, the approximation error never really vanishes. 

Now, what happens when the model has to predict a value for $x^t_i$ such that either $x_i$ or the projection lies beyond the boundary values it has learned in training?  The boundary values and the parameter $\alpha_M$ from Observation \ref{obs:boundary} define the limits of what a "close" element is.  If $\pi(x_i) = y_i$ is not too far from $[B^{-}, B^{+}]$,then the closest element the model can find is the boundary value or something near it.  Far away elements are $Ker(\pi)$ and so sent to $0$ or some nearby value.  Observation \ref{obs:boundary} establishes that the projection is not linear over intervals larger than $[B^-, B+]$, and that's what our algorithm also predicts. % the model no longer finds a close element from which to calculate the projection and then picks a random projection in memory and returns a random element from $[B^{-}, B^{+}]$.

%This is, we conjecture, because they provide more memory for stocking nearby sequences from which the model can draw to optimize $\pi$ (see Appendix F).  A sequence $\vec{y}$ stored  in $V^{h,l}$ that is far away from $\vec{x}$ will have the effect that $V^{h,l}(\vec{x})$ is very small or 0; if $|\vec{y} = \vec{x}| < \alpha_M$, $\alpha_M$ is $M$'s model parameter from Observation \ref{obs:boundary}), then $V^{h,l}(\vec{x})$ will be high. 

So, even though memory is a crucial component of ICL$_1$, it's not everything.  The autoregressive component of our algorithm above resembles basic causal language modeling, but it incorporates a basic reasoning element of closeness.  The closer a sequence stored in memory of a model M $\vec{z}$ is to the target sequence, the higher the probability that $\pi_M(x_n) = ay^{\vec{z}}_n$, for $a < \alpha_M$.   %For a model $M$ all non $\alpha_M$-close sequences will be in $Ker(\pi)$.   

\section{Conclusion}
While transformers are theoretically capable of learning linear regression to perform ICL ${\cal L}$, we have demonstrated they do not do this in practice. Our models achieve strong statistical results comparable to state-of-the-art algorithms when training and test distributions are the same.  However, we aimed to assess whether our models truly learn the underlying class representation, and they do not.  All tested models failed to robustly learn the class form $f(x) = ax + b$ of linear functions on both noisy and non-noisy data. Had models used linear regression on our clean data, they would have exactly inferred the class form of linear functions and correctly estimated the parameters $a,b$. 

We also found that attention layers were both necessary and sometimes sufficient for good ICL performance.  More layers and more heads improved performance.  Generalization ability improved as perhaps expected when models were trained on distributions with larger variance.  

We also analyzed model errors and emphasized the importance of boundary values that define intervals within which the vast majority of all of values of the training sequences occur.    % and still perform almost optimally on the original target distribution (say $\mathcal{N}(0,1)$ with more layers.  But in addition, we saw that models trained on a distribution $D$ like $U(-10,10)$ that includes examples of functions $f$ and inputs $x_i$ not likely to be in $\mathcal{N}(0,1)$ or not at all present in $U(-1,1)$ still performs well when tested on $U(-1,1)$ or $\mathcal{N}(0,1)$ but generalizes far better than models trained only on $\mathcal{N}(0,1)$.  
End users, who typically lack knowledge about the training data distribution for the models they use, need to know about model sensitivity to training and testing regimes. Otherwise, they may unwittingly employ ICL on tasks significantly beyond the original training domain with a consequent substantial loss of accuracy. This limitation will affect larger models working on different tasks.  %We hope our work will inspire further research into what transformer-based models are actually doing on ICL tasks.

Finally, we have offered a hypothesis about how models learn.
  Our hypothesis $H$ explains our observations, but we have not
  fully shown that transformer models actually implement this hypothesis.  In future work, we will attempt to isolate various stages of the algorithm by probing what models do at various layers, following \cite{wang2022interpretability,nanda2023progress}.  %Mechanistic interpretability, however, is much more difficult in purely numerical data and predictions than for language models that output language sequences, because the latter are at least to some extent {\em per se} interpretable.  

\begin{credits}
\subsubsection{\ackname}
We gratefully acknowledge the support of the EU grant TUPLES, the grants SARER, Summ-RE (ANR-20-CE23-0017), and the AI Cluster ANITI (ANR-19-PI3A-0004). This work used the HPC resources from CALMIP (Grant 2016-P23060). 
\end{credits}

\bibliographystyle{splncs04}
\bibliography{biblio}

\appendix

\section{Training details}
\label{sec:appendixA}
%\large{\bf Appendix A: Training details} 

\textbf{Additional training information:} We use the Adam optimizer \cite{diederik2014adam} , and a learning rate of $10^{-4}$ for all models.\\
\textbf{Computational resources:} We used Nvidia A-100 GPUs and Nvidia Volta (V100 - 7,8 Tflops DP) for every training involved in these experiments. \\

\section{Error progression for models trained on $\mathcal{N}(0,1)$ distributions tested on $\mathcal{N}(0,\sigma)$}
\label{sec:appendixB}
\begin{table*}[!h]
\small{
\begin{tabular}{l|l|l|l|l|l|l|l|l|l|l}
 \hline
  models \ $\backslash$ \ $\sigma$ & 1 & 2 & 3 & 4 & 5 & 6 & 7 & 8 & 9 & 10 \\ 
 \hline\hline
 1L1AH $d_{embedding}$=64  & 0.1 & 0.8 & 5.1 & 13.1 & 26.9 & 39.7 & 53.0 & 84.8 & 120.0 & 153.2 \\

 1L2AH $d_{embedding}$=64  & 0.1 & 0.8 & 5.3 & 14.4 & 29.8 & 41.1 & 55.0 & 93.8 & 120.4 & 159.2 \\

 1L4AH $d_{embedding}$=64  & 0.0 & 0.2 & 2.7 & 8.7 & 19.9 & 32.0 & 42.8 & 64.5 & 92.3 & 131.2 \\
 \hline
 2L1AH $d_{embedding}$=64  & 0.0 & 0.1 & 2.0 & 4.9 & 13.7 & 27.0 & 36.1 & 64.9 & 99.0 & 134.0 \\

 2L2AH $d_{embedding}$=64  & 0.0 & 0.0 & 1.6 & 3.2 & 9.3 & 25.5 & 32.0 & 61.1 & 92.9 & 127.8 \\

 2L4AH $d_{embedding}$=64  & 0.0 & 0.0 & 0.9 & 2.6 & 7.5 & 19.3 & 27.3 & 51.8 & 90.2 & 119.4 \\
 \hline
 3L1AH $d_{embedding}$=64  & 0.0 & 0.0 & 0.9 & 3.0 & 8.2 & 16.8 & 24.4 & 48.4 & 76.7 & 113.2 \\

 3L2AH $d_{embedding}$=64  & 0.0 & 0.0 & 0.7 & 2.3 & 6.5 & 15.9 & 22.5 & 43.1 & 74.0 & 102.5 \\

 3L4AH $d_{embedding}$=64  & 0.0 & 0.0 & 0.6 & 1.9 & 5.5 & 13.8 & 20.4 & 42.2 & 70.3 & 100.4 \\
 \hline
 6L4AH $d_{embedding}$=64  & 0.0 & 0.0 & 0.5 & 1.6 & 4.6 & 11.6 & 16.8 & 33.7 & 58.3 & 87.9 \\
 \hline
12L8AH $d_{embedding}$=256  & 0.0 & 0.0 & 0.3 & 1.1 & 2.9 & 7.9 & 11.9 & 28.3 & 46.9 & 73.5 \\ [1ex] 
 \hline
 18L8AH $d_{embedding}$=256  & 0.0 & 0.0& 0.2& 1.1& 2.8& 7.1& 10.3& 22.9& 40.3& 64.6 \\ [1ex] 
 \hline
 3NN  & 0.03 & 0.14 & 0.27 & 0.66 & 1.09 & 1.32 & 1.75 & 2.45 & 2.95 & 4.01 \\ [1ex] 
 \hline
 $REF_{D^t_F,D^t_I}: y=0$   & 2.19 & 7.05 & 19.22 & 33.94 & 52.23 & 73.08 & 86.02 & 127.43 & 165.27 & 199.31 \\ [1ex] 
 \hline
\end{tabular}
}

\caption{Comparison to show the evolution of squared $\epsilon$ type error depending on the distribution according to which we take the parameters, without taking into account the error of the prediction of the first and second prompts. $D_i^t = \mathcal{N}(0,1)$. 3NN refers to \cite{olsson:etal:2022}'s method which generates values through the average of the 3 nearest neighbors}
\label{table:3}
\end{table*}
%\guilhem{We should here reference the table \ref{table:3}}
%The table of this appendix is Table \ref{table:3}. \\
When $D_I = D_F = \mathcal{N}(0,\sigma)$ there is  for $x \in  \mathcal{N}(0,\sigma)$  an over %68\% chance that a function chosen for train $f$ will have $f(x)\in [-\sigma, \sigma]$ and over a
 85\% chance of  $f(x) \in [-4 \sigma^{2} -  2 \sigma, 4 \sigma^{2} + 2 \sigma]$ and a  95\% chance $f(x) \in [-2\sigma, 2\sigma]$. So a model with $\sigma = 1$ $D_F = D_I = \mathcal{N}(0,1)$ has seen sequences of values for $f$ with $f(x) \in [-2,2]$ more than 95\% of the time. \\

\section{Error values for models on different distributions}
\label{sec:appendixC}

The results are in Table \ref{table:1}

\begin{table*}
\small{
\begin{tabular}{l l l l l l l l l l l}
 \hline
 models \ $\backslash$ \ $\sigma$ & 1 & 2 & 3 & 4 & 5 & 6 & 7 & 8 & 9 & 10 \\ 
 \hline\hline
 $3L4AH_N$, $d_{emb}=64$   & 0.0 & 0.0 & 0.22 & 0.4 & 1.73 & 6.56 & 8.56 & 20.44 & 39.73 & 53.93 \\

 $3L4AH_B$, $d_{emb}=64$   & 0.03 & 0.15 & 0.53 & 1.32 & 2.74 & 3.91 & 5.52 & 10.22 & 13.86 & 22.72 \\
 
  $3L4AH_U$, $d_{emb}=64$   &  0.02 & 0.03 & 0.13 & 0.36 & 0.84 & 1.79 & 2.54 & 7.06 & 11.38 & 17.75 \\ [1ex] 
 \hline\hline
 $6L4AH_N$, $d_{emb}=64$   & 0.0 & 0.0 & 0.2 & 0.38 & 1.58 & 5.72 & 7.99 & 15.53 & 32.96 & 50.35 \\

  $6L4AH_B$, $d_{emb}=64$   & 0.01 & 0.04 & 0.23 & 0.44 & 1.19 & 2.15 & 3.08 & 4.8 & 9.98 & 18.01 \\

   $6L4AH_U$, $d_{emb}=64$   & 0.02 & 0.04 & 0.11 & 0.24 & 0.57 & 1.36 & 1.82 & 4.62 & 10.23 & 15.07 \\[1ex] 
 \hline\hline
 $12L8AH_N$, $d_{emb}=256$  & 0.0 & 0.0 & 0.32 & 1.34 & 3.14 & 8.8 & 12.13 & 30.14 & 49.37 & 73.93 \\  

  \textbf{sorted $12L8AH_N$}  & 0.0 & 0.01 & 0.32 & 1.63 & 3.69 & 8.39 & 10.06 & 27.11 & 43.23 & 58.56 \\  

  \hline
 $12L8AH_B$, $d_{emb}=256$  & 0.0 & 0.01 & 0.08 & 0.29 & 0.78 & 2.23 & 3.66 & 9.04 & 18.68 & 30.23 \\ 

 \textbf{sorted $12L8AH_B$}  & 0.01 & 0.03 & 0.18 & 0.25 & 0.74 & 2.27 & 2.62 & 6.87 & 13.73 & 20.8 \\ 

  \hline
 $12L8AH_U$, $d_{emb}=256$  & 0.0 & 0.01 & 0.13 & 0.71 & 1.92 & 6.78 & 10.92 & 27.91 & 38.75 & 64.39 \\ [1ex] 

 \textbf{sorted $12L8AH_U$}   & 0.01 & 0.01 & 0.13 & 0.75 & 2.12 & 6.18 & 10.5 & 26.8 & 36.3 & 53.48 \\ [1ex] 
 \hline\hline
  \textbf{$REF_{D^t_F,D^t_I}$: y=0}   & 1.52 & 4.43 & 13.55 & 19.94 & 30.81 & 44.75 & 52.71 & 76.11 & 105.43 & 128.52 \\ [1ex] 
 \hline
\end{tabular}
}

\caption{Comparison showing the evolution of squared errors for models trained on different distributions; index N: $D_F = \mathcal{N}(0,1)$, B $D_F = 0.5{\cal N}(-1,1) + 0.5 {\cal N}(1,1)$ and U $D_F = {\cal U}(-5,5)$. We show error rates for models prompted without and with the natural ordering on the prompts [sorted], for the large model size. $D^t_i = {\cal U}(-1,1)$ and  $D^t_F=\mathcal{N}(0,\sigma)$}
\label{table:1}
\end{table*}

\newpage

\section{Results on attention layer only models}
\label{sec:appendixD}

\begin{figure}[!h]
\center 
\includegraphics[width=8cm, alt={Plot showing the evolution of error rates for models composed only of attention layers. The models include. All models use embeddings of size 64, except 2AL32AH, which uses size 256.}]{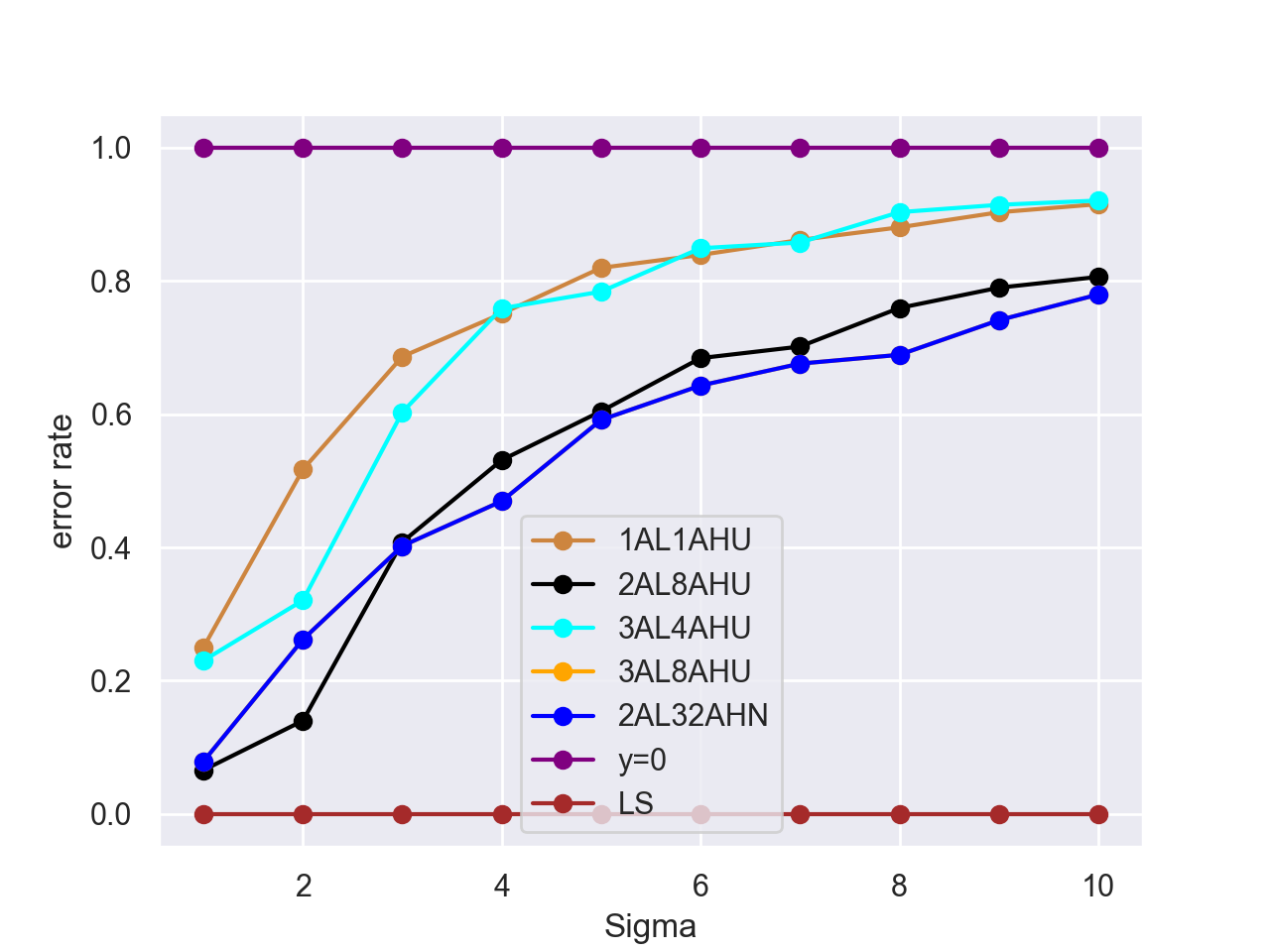}
\caption{Evolution of error rates for models with attention layers only. We give figures for a model with only 1 attention layer/1AH (1AL1AH) two 2-attention layer only models  (2AL8AH, 2AL32AH), two 3 attention layer only model  (3AL4AH,3AL8AH), and 12 attention layer model only (12L8AH). $D_I=D_F={\cal U}(-1,1)$, $D^t_i = {\cal U}(-1,1)$ and  $D^t_F=\mathcal{N}(0,\sigma)$.  All models have embeddings of size 64, except $2AL32AH$ has size 256.
\label{progressive-lossAH}}
\end{figure}
 The 12L8AH attention-only model exhibited strong performance, closely matching that of the 12L8AH model with an MLP. The large two-attention-only layer model with 32 attention heads demonstrated greater robustness compared to the full transformer model (which includes an MLP) with either a single layer and one attention head or two attention heads. (See Table \ref{table:3}). A single AL model had only a very limited ICL generalization capability beyond testing on $D^t_F = \mathcal{N}(0,1)$. %, but it did better than a 12 layer MLP, which showed no ICL capability.% probably because the method of training on the predict next token format is not suitable for models without attention heads.   A
 Tables \ref{table:2} and \ref{table:4}  in  Appendix and Figure \ref{progressive-lossAH}  give details of various AL models on normal and uniform distributions. 
%\newpage
%\large{\bf Appendix F: The model searches for a sequence close to the input sequence.}

%\newpage
\section{Plots for boundary values with $\mathcal{N}(0,1)$ and ${\cal U}(-5,5)$}
\label{sec:appendixE}
%\large{\bf Appendix D: Plots for boundary values with $\mathcal{N}(0,1)$ and $U(-5,5)$} \\

\begin{figure}[ht!]
\centering
\includegraphics[width=0.4\textwidth, alt={Plot showing the predictions of the 12L8AH model trained on data from a standard normal distribution, when evaluated on the function f of x equals 30 times x plus 30.}]{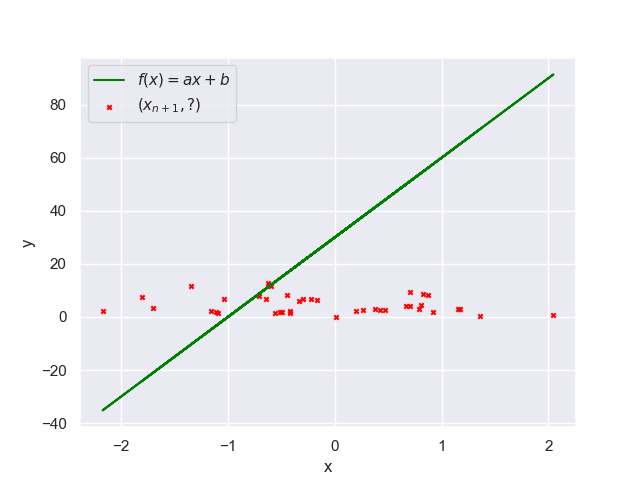}
\includegraphics[width=0.4\textwidth,  alt={Plot showing the squared errors of the 12L8AH model trained on data from a standard normal distribution, when evaluated on the function f of x equals 30 times x plus 30.}]{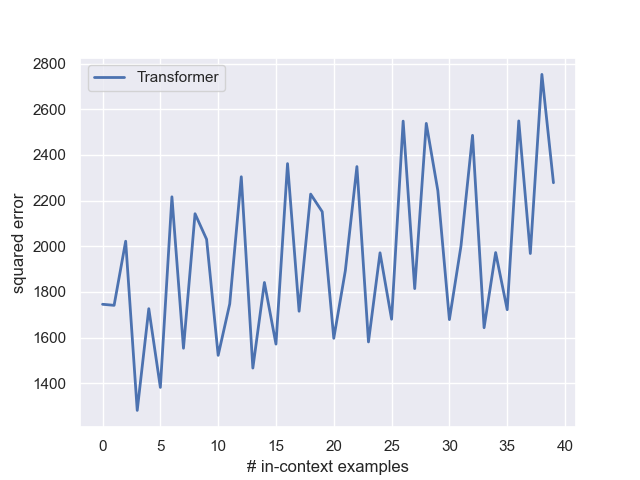} \\
\includegraphics[width=0.4\textwidth, alt={Plot showing the predictions of the 2L32AH attention-only model, using an embedding size of 256, on the function f of x equals x.}]{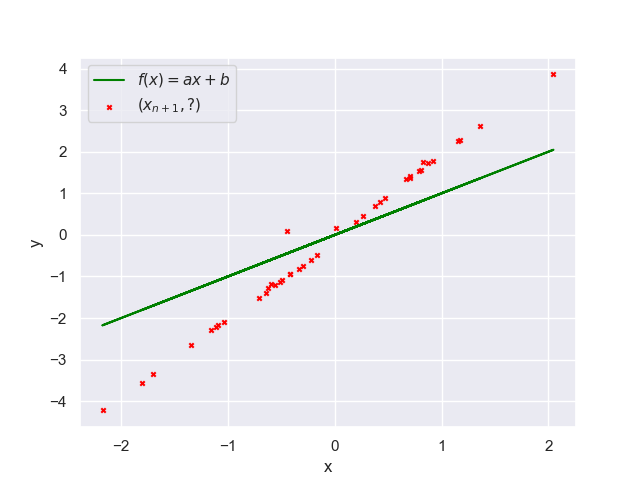}
\includegraphics[width=0.4\textwidth, alt={Plot showing the predictions of the 2L32AH attention-only model, using an embedding size of 256, on the function f of x equals 15x. }]{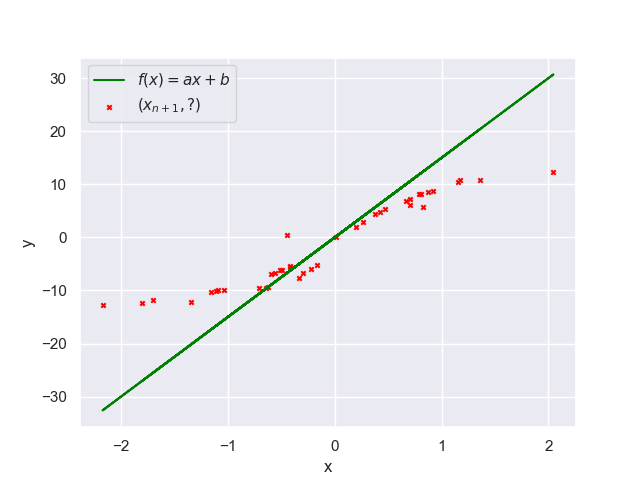}
\caption{
The first and second plots show respectively, predictions for the 12L8AH model trained on $\mathcal{N}(0,1)$ and its error evolution over number of prompts, both for $f(x) = 30x + 30$. The third and fourth plots show predictions of $f(x) = x$ and $f(x) = 15x$ for 2L32AH attention only model with $d_{embedding}=256$  \label{big30}}

\end{figure}

%The figures of this appendix are: Figures \ref{40x+40} \ref{big30} \ref{fig:boundary} \ref{relu_shape2}.

%Here we add some more examples
\begin{figure}[ht!] 
\centering
\includegraphics[width=6cm, alt={Plot showing boundary values with the predictions of the 3L4AH model on the function f of x equals 9.4 times x. The model is trained and tested on input and target data sampled uniformly from -5 to 5. }]{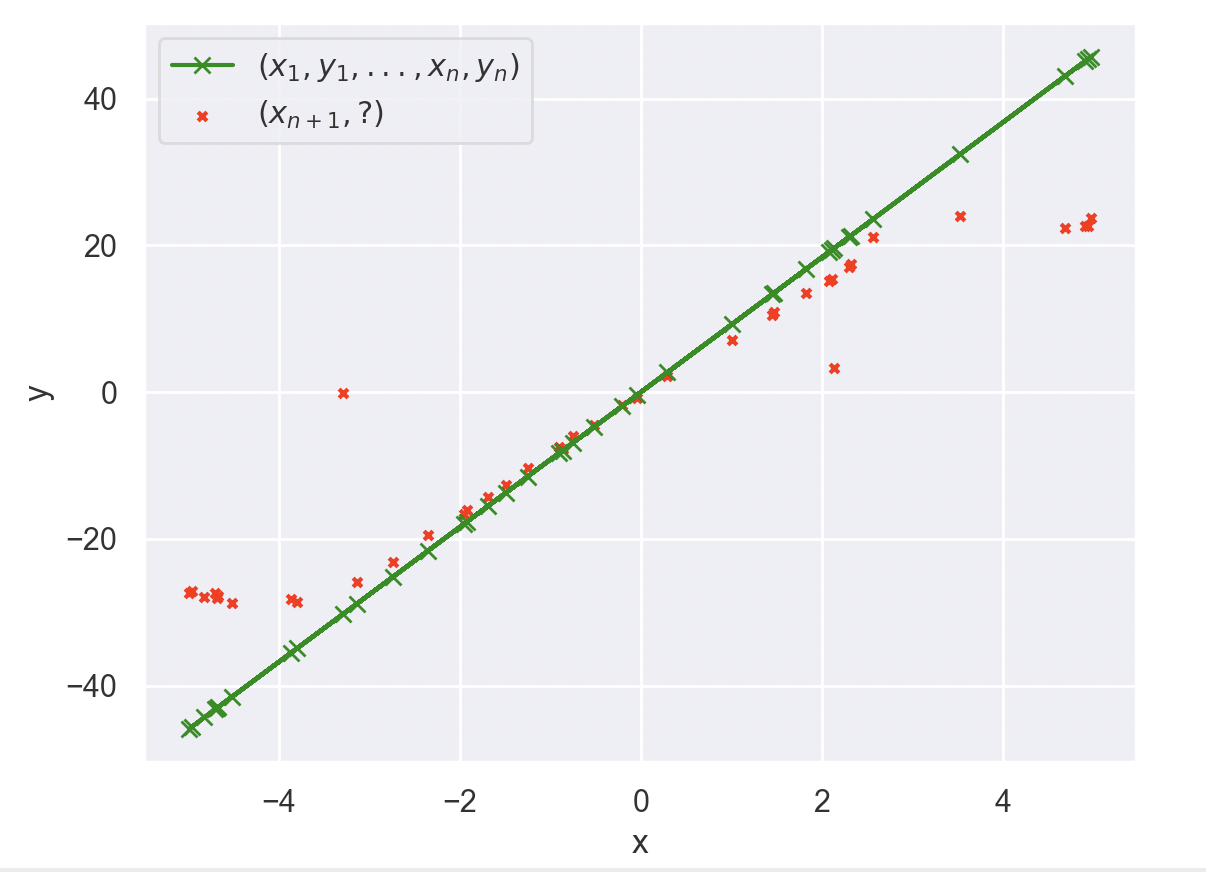}
\includegraphics[width=6cm, alt= {Plot showing boundary values with the predictions of the 6L4AH model on the function f of x equals 9.4 times x. The model is trained and tested on input and target data sampled uniformly from -5 to 5. }]{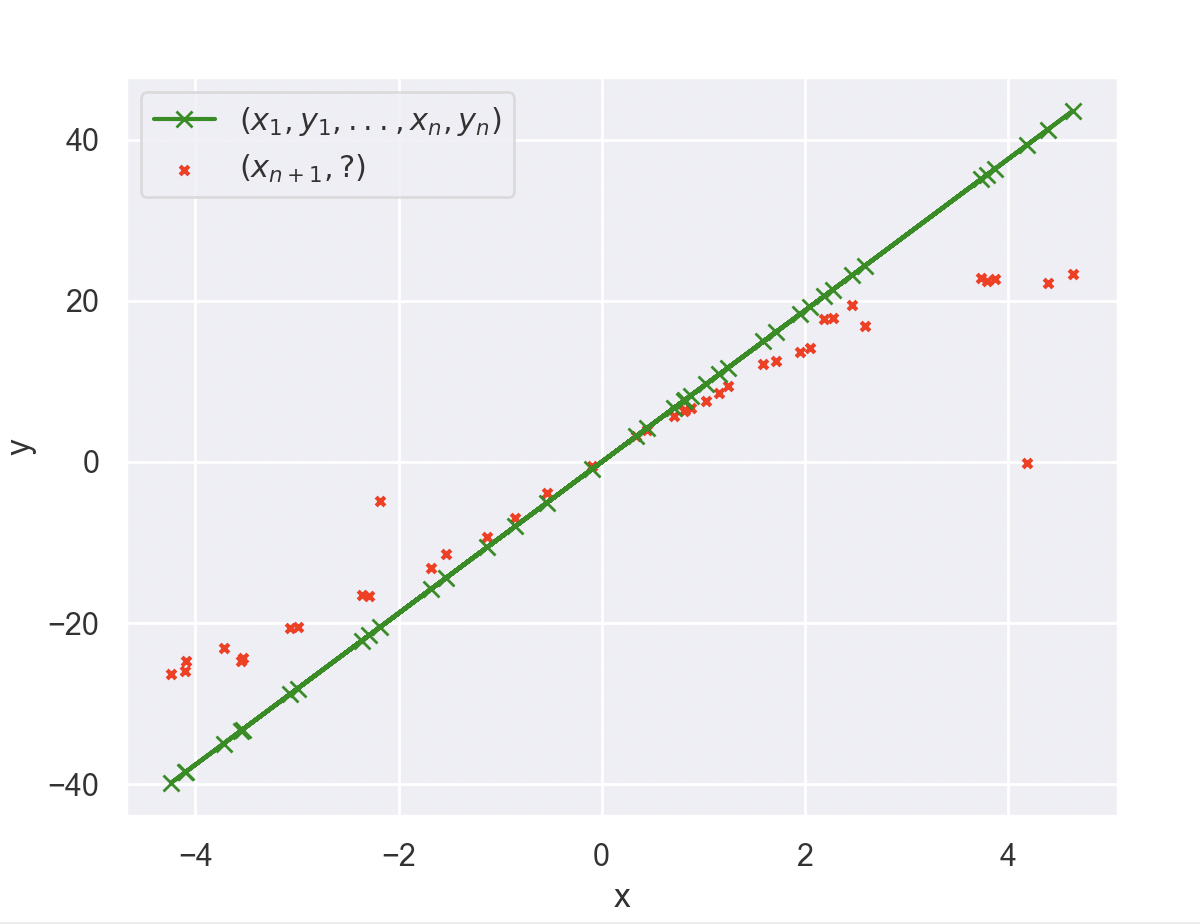}\label{relu-shape3}\\
\caption{Boundary values: Plots for $f(x) = 9.4x$ for models 3L4AH and 6L4AH, $D_I = D_F = D_I^t = D_F^t = {\cal U}(-5,5)$\label{relu_shape2}}
\end{figure}

\begin{figure}[ht!]  
\centering
\includegraphics[width=6cm, alt={Plot showing the predictions of the 12L8AH model on the function f of x equals 10 times x. }]{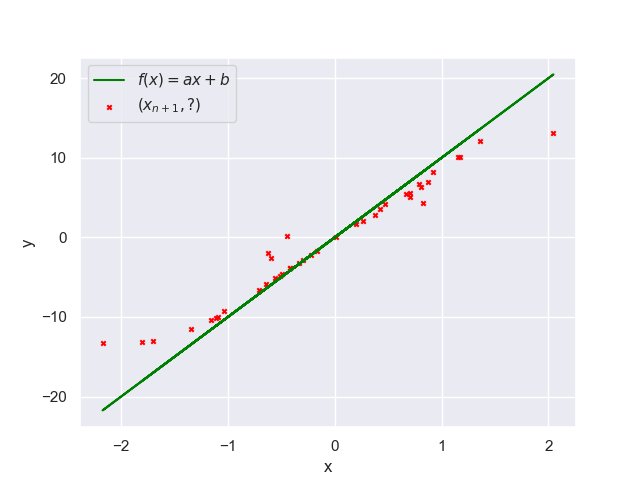}
\includegraphics[width=6cm, alt={Plot showing the predictions of the 6L4AH model on the function f of x equals 10 times x. }]{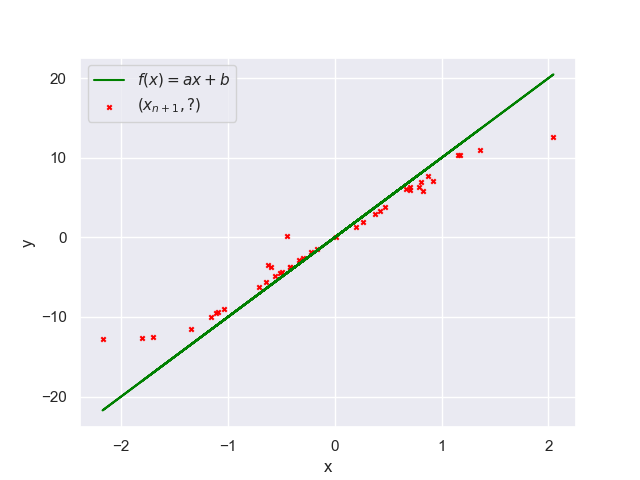}
\caption{Plots for $f(x) = 10x$ by a 12L8AH model and by a 6L4AH model.} \label{fig:boundary}
\end{figure}

\newpage
\section{Example of boundary values for attention only models}
\label{sec:appendixF}
%The tables of this appendix are \ref{table:2} \ref{table:4}.% with the figure \ref{fig:relu-ahonly}.
%\large{\bf Appendix E: Example of boundary values for attention only models}
\begin{figure}[ht!]
\center \includegraphics[width=5.5cm]{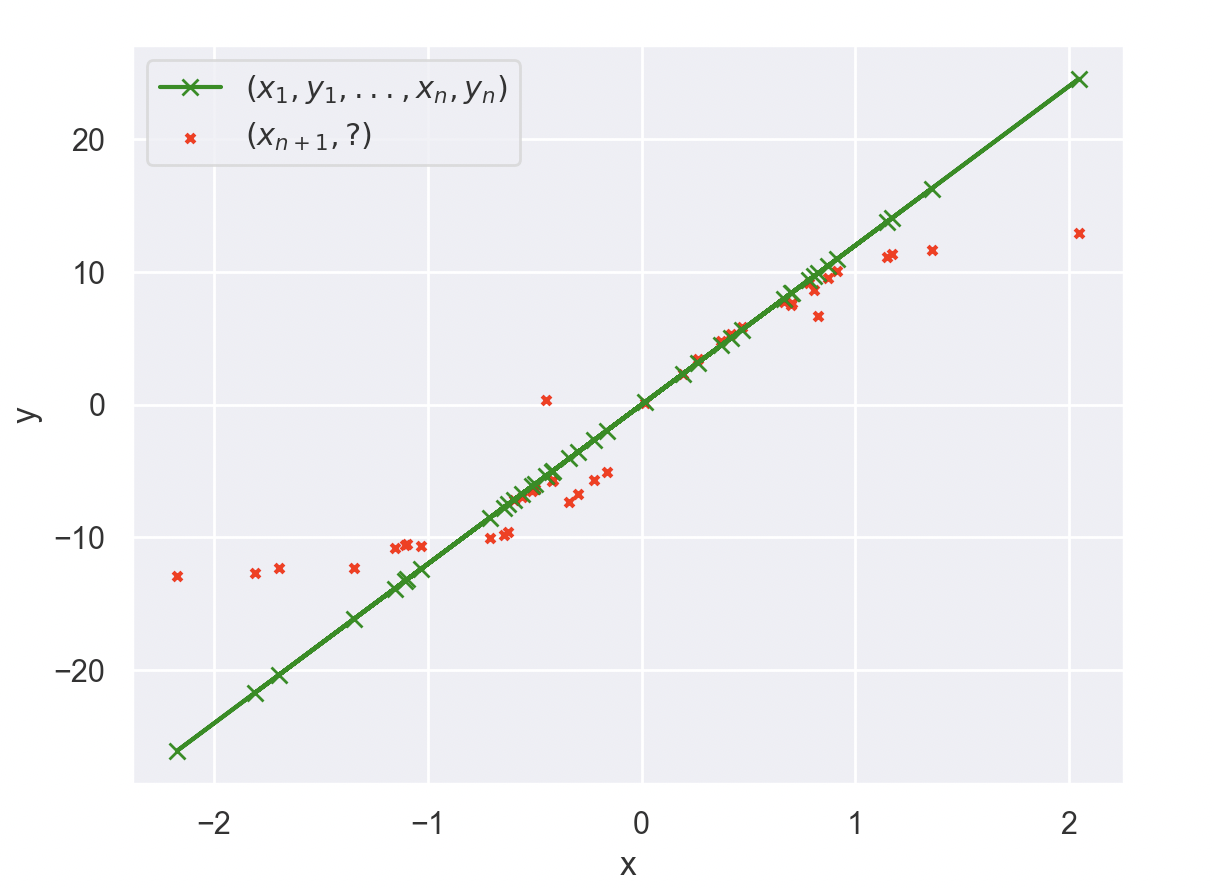}
\caption{Boundary values for 2L32ah attention only model, with $d_{embedding}= 256$ to ICL the function $f(x) = 12x$ } \label{fig:relu-ahonly}
\end{figure} 

\begin{table*}[!h]
\small{
\begin{tabular}{l|l|l|l|l|l|l|l|l|l|l}
 \hline
  models \ $\backslash$ \ $\sigma$ & 1 & 2 & 3 & 4 & 5 & 6 & 7 & 8 & 9 & 10 \\ 
\hline\hline
 %$3L4AH_N$   & 0.0 & 0.0 & 0.22 & 0.4 & 1.73 & 6.56 & 8.56 & 20.44 & 39.73 & 53.93 \\
% \hline
% $3L4AH_B$,   & 0.03 & 0.15 & 0.53 & 1.32 & 2.74 & 3.91 & 5.52 & 10.22 & 13.86 & 22.72 \\
% \hline
%  $3L4AH_U$   &  0.02 & 0.03 & 0.13 & 0.36 & 0.84 & 1.79 & 2.54 & 7.06 & 11.38 & 17.75 \\ [1ex] 
% \hline\hline
 $1AL1AH_{U}$   & 0.38 & 2.29 & 9.3 & 14.97 & 25.25 & 37.54 & 45.4 & 67.0 & 95.19 & 117.6 \\ [1ex] 

 $2AL8AH_{U}$   &  0.1 & 0.62 & 5.53 & 10.59 & 18.62 & 30.61 & 36.97 & 57.79 & 83.26 & 103.58 \\ [1ex] 

% $2Al32AH_U$ &  0.86 & 1.61 & 3.53& 10.95& 22.43 & 35.3 & 46.98 & 67.12 & 104.83 & 135.21 \\
% \hline
 $3AL4AH_{U}$   &  0.35 & 1.42 & 8.17 & 15.13 & 24.15 & 37.99 & 45.2 & 68.73 & 96.37 & 118.3 \\ 

 $3AL8AH_{U}$   &  0.12 & 1.16 & 5.45 & 9.36 & 18.22 & 28.77 & 35.62 & 52.44 & 78.12 & 100.18 \\ [1ex] 

  $2Al32AH_N$ & 0.06 & 0.91 & 5.96 & 10.43 & 18.96 & 30.11 & 36.77 & 55.59 & 81.66 & 103.17\\
$12Al8AH_N$ &0.0& 0.0& 0.41& 1.70& 3.92& 10.40& 14.04& 30.20& 52.69& 79.13\\
 \hline\hline
 $REF_{D^t_F,D^t_I}: y=0$   &  1.52 & 4.43 & 13.55 & 19.94 & 30.81 & 44.75 & 52.71 & 76.11 & 105.43 & 128.52 \\ [1ex] 
 \hline
%  \hline\hline
%  $2Al32AH_N$ &1.17 & 2.64& 3.47& 5.01& 7.88& 16.85& 24.1& 40.98& 66.04& 95.03\\
%\hline
\end{tabular}
}
\caption{Comparison showing the evolution of squared errors for  models with attention layers only. We give figures for a model with only 1 attention layer/1AH (1AL1AH) two 2-attention layer only models  (2AL8AH, 2AL32AH) and two 3 attention layer only model  (3AL4AH,3AL8AH). $D_I=D_F={\cal U}(-1,1)$, $D^t_i = {\cal U}(-1,1)$ and  $D^t_F=\mathcal{N}(0,\sigma)$.  All models have embeddings of size 64, except 2AL32AH has size 256. }
\label{table:2}
\end{table*}

\begin{table*}[!h]
\small{
\begin{tabular}{l|l|l|l|l|l|l|l|l|l|l}
 \hline
  models \ $\backslash$ \ $\sigma$ & 1 & 2 & 3 & 4 & 5 & 6 & 7 & 8 & 9 & 10 \\ 
 \hline\hline
 $1L1AH_N$ $d_{embedding}$=64  & 48.8 & 57.62 & 73.48 & 84.51 & 116.63 & 129.52 & 142.34 & 177.69 & 191.05 & 246.43 \\

 $2L8AH_N$ $d_{embedding}$=64  & 2.24 &4.81 & 5.8 & 7.19 & 10.01 & 19.04 & 30.22 & 38.03 & 73.32 & 118.89 \\

$2L32AH_N$ $d_{embedding}$=256  & 1.17 & 2.64 & 3.47 & 5.01 & 7.88 & 16.85 & 24.1 & 40.98 & 66.04 & 95.03 \\ [1ex] 
 \hline \hline
 $REF_{D^t_F,D^t_I}: y=0$   & 2.19 & 7.05 & 19.22 & 33.94 & 52.23 & 73.08 & 86.02 & 127.43 & 165.27 & 199.31 \\ [1ex] 
 \hline
\end{tabular}
}

\caption{Comparison to show the evolution of squared $\epsilon$ type error depending on the distribution according to which we take the parameters, without taking into account the error of the prediction of the first and second prompts. $D_F=D_I=D_i^t = \mathcal{N}(0,1)$ for models with attention ONLY
}
\label{table:4}
\end{table*}

\newpage
\section{How models calculate values versus \cite{olsson:etal:2022}'s proposal}
\label{sec:appendixH}
Models don't correct their previous predictions each time they predict a new one.  That is, the autoregressively predicted values remain unchanged as more examples are provided. The example below demonstrates this behavior: the model generates four values after three are provided, and then generates the same four values when an additional (fourth) example is given.\\
In this example we take, $f(x)=x$ for $x \in \{0,0.1,...,0.5\}$ \\
In the first line we give as prompt $(0,0,0.1,0.1,0.2,0.2,0.3,0.3,0.4,?)$ and in the second $(0,0,0.1,0.1,0.2,0.2,0.3,0.3,0.4,0.4,0.5,?)$ \\
Below are the values predicted by the model.
\begin{tcolorbox}[colback=green!5!white,colframe=green!75!black]
-0.0052 | 0.1001 | 0.2961 | {\color{cyan}0.4123}
  \tcblower
-0.0052 | 0.1001 | 0.2961 | 0.4123 | {\color{cyan}0.5237}
\end{tcolorbox}

\cite{olsson:etal:2022} suggests that models average over three closest neighbors.  If we do this, the value predicted will be $0.15$ for the first input, which is far from $0.4$ and $0.3$ instead of $0.5$ for the second example.  The model's method is clearly superior. \\

Here is another example showing the limits limits of \cite{olsson:etal:2022}'s proposition of what the model might be learning.
Let's take $f(x)=x$ for $x \in [ x_1=0.0144, x_2=-0.4471, x_3= -0.6244, x_4=-0.5978]$.  Consider the prompt $$(x_1,f(x_1),...,x_3,f(x_3),x_4,?)$$ The trained model predicts $-0.5951$ but \cite{olsson:etal:2022}'s proposition returns $-0.3524$. The accuracy of the proposal clearly depends on the sample we got and if it has values really near to the target or not.

Call our method $H$ for computing $y_n$ given $(x_1, y_1,..., x_n)$ and call a simple averaging method like \cite{olsson:etal:2022}'s $A$.
\begin{proposition}  $P(H(\vec{x}) < f(x_n) +\epsilon) >>  P(A(\vec{x}) < f(x_n) +\epsilon)$. 
\end{proposition}
Consider the uniform distribution {\cal U}(-1,1). $$P(x_i - \epsilon \leq X \leq x_i + \epsilon) = \int_{x_i - \epsilon}^{x_i + \epsilon} \frac{1}{1-(-1)} dx = \epsilon $$
However, as $H$ refines the projection $\pi$, $P(\pi(x_i) < f(x_n) + \epsilon) = i^m\times\epsilon^n$, where $i > 0, m,n < 41$.
On the other hand, $P(A(\vec{x}) < f(x_n) +\epsilon) \approx 0$).

%\large{\bf Appendix D:Plots for ICL over number of prompts}
%\begin{figure}[!ht] 
%\caption{ Plot of ICL over number of prompts for $f(x) = x$ with $D_F=D_I=D_I^t= U(-5,5)$ for the model 12L8AH\label{p2>p1}}
%\end{figure}
%\newpage

%\newpage

%\newpage

\hidden{
\begin{table*}
\small{
\begin{tabular}{l l l l l l l l l l l}
 \hline
 models \ $\backslash$ \ $\sigma$ & 1 & 2 & 3 & 4 & 5 & 6 & 7 & 8 & 9 & 10 \\ 
 \hline\hline
 $12L8AH$, $d_{emb}=256$   & 65098.6& 44032.5& 33789.9& 26700.7& 20029.1& 16505.8& 15452.8& 16672.8& 15524.01& 14787.2 \\
 \hline
\end{tabular}
}

\caption{Comparison showing the evolution of squared errors for models trained on $D_F = D_I = \mathcal{N}(0,100)$  $D^t_i = {\cal U}(-1,1)$ and tested on $D^t_F=\mathcal{N}(0,\sigma)$ and $D^t_I=\mathcal{N}(0,1)$}
\label{table:10}
\end{table*}
}

\end{document}